\documentclass[journal]{IEEEtran}
\usepackage{graphicx}
\usepackage{epstopdf}
\usepackage{color}
\usepackage[table,xcdraw]{xcolor}
\usepackage{multirow}
\usepackage[pagebackref=false,colorlinks=true,urlcolor = blue,bookmarks=false]{hyperref}
\usepackage{amsmath}
\usepackage{amssymb}
\usepackage{times}
\usepackage{epsfig}
\usepackage{graphicx}
\usepackage{amsmath}
\usepackage{amssymb}
\usepackage[table,xcdraw]{xcolor}
\usepackage{multirow}
\usepackage{booktabs}
\usepackage{bbding}
\usepackage{pifont}
\usepackage[figuresright]{rotating}
\usepackage{color}
\usepackage{diagbox}
\usepackage[numbers,sort&compress]{natbib}
\usepackage[switch]{lineno}
\begin{document}

\title{LasHeR: A Large-scale High-diversity Benchmark for RGBT Tracking}
\author{Chenglong Li, Wanlin Xue, Yaqing Jia, Zhichen Qu, Bin Luo, Jin Tang, and Dengdi Sun~\thanks{This work is partly supported by NSFC Key projects in International (regional) cooperation and exchanges (No. 61860206004), National Natural Science Foundation of China (No. 61976003, 62076003, 62076005), Natural Science Foundation of Anhui Province (2008085MF191), and Open Projects Program of National Laboratory of Pattern Recognition. (\emph{Corresponding author: Dengdi Sun})}
\thanks{C. Li and D. Sun are with Information Materials and Intelligent Sensing Laboratory of Anhui Province, Anhui Provincial Key Laboratory of Multimodal Cognitive Computation, School of Artificial Intelligence, Anhui University, Hefei 230601, China. (Email: lcl1314@foxmail.com, sundengdi@163.com).}
\thanks{W. Xue, Y. Jia, Z. Qu, B. Luo and J. Tang are with Information Materials and Intelligent Sensing Laboratory of Anhui Province, Key Lab of Intelligent Computing and Signal Processing of Ministry of Education, Anhui Provincial Key Laboratory of Multimodal Cognitive Computation, School of Computer Science and Technology, Anhui University, Hefei 230601, China. (Email: xwl\_0929@163.com, yqjia1024@163.com, E01814248@stu.ahu.edu.cn, luobin@ahu.edu.cn, tangjin@ahu.edu.cn).}}

%\author{A,}

\markboth{IEEE TRANSACTIONS ON IMAGE PROCESSING}%
{Shell \MakeLowercase{\textit{et al.}}: Bare Demo of IEEEtran.cls for IEEE Journals}

\maketitle
\begin{abstract}
RGBT tracking receives a surge of interest in the computer vision community, but this research field lacks a large-scale and high-diversity benchmark dataset, which is essential for both the training of deep RGBT trackers and the comprehensive evaluation of RGBT tracking methods. To this end, we present a \emph{La}rge-\emph{s}cale \emph{H}igh-diversity b\emph{e}nchmark for short-term \emph{R}GBT tracking (LasHeR) in this work. LasHeR consists of 1224 visible and thermal infrared video pairs with more than 730K frame pairs in total. Each frame pair is spatially aligned and manually annotated with a bounding box, making the dataset well and densely annotated. LasHeR is highly diverse capturing from a broad range of object categories, camera viewpoints, scene complexities and environmental factors across seasons, weathers, day and night. We conduct a comprehensive performance evaluation of 12 RGBT tracking algorithms on the LasHeR dataset and present detailed analysis. In addition, we release the unaligned version of LasHeR to attract the research interest for alignment-free RGBT tracking, which is a more practical task in real-world applications. The datasets and evaluation protocols are available at: https://github.com/mmic-lcl/Datasets-and-benchmark-code.
\end{abstract}


\section{Introduction}

\IEEEPARstart{V}{isible} and thermal infrared information strongly complement each other and contribute to visual tracking in different aspects. On one hand, thermal infrared radiation(3-15$\mu$m) are not affected by lighting variations and have a strong penetration ability to some particulate matters (e.g., smog and fog). On the other hand, visible data are more effective in discriminating targets from other objects or background when they are in thermal crossover. Therefore, RGBT tracking, aiming to locate targets with a bounding box using visible and thermal information given the initial state, has received a surge of interest in the computer vision community. 
\begin{figure}[t]
\begin{center}
   \includegraphics[width=0.9\linewidth]{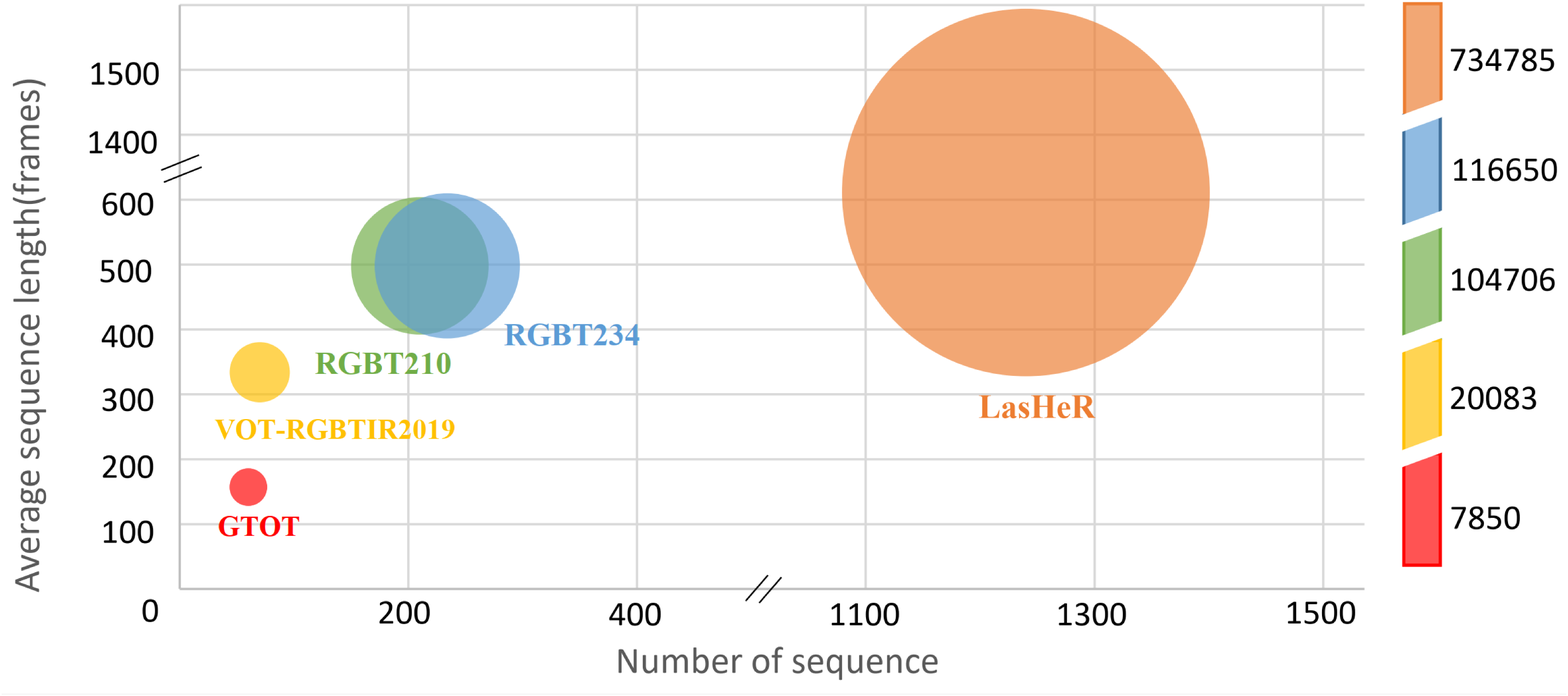}
\end{center}
   \caption{Comparison of the proposed LasHeR with existing RGBT tracking benchmark datasets, including GTOT~\cite{Li16tip}, VOT-RGBTIR2019~\cite{Kristan2019a}, RGBT210~\cite{Li17acmmm} and RGBT234~\cite{Li2018pr}. The circle diameter is in proportion to the number of frames of a benchmark dataset.}
\label{fig:bubble}
\vspace{-0.3cm}
\end{figure}
\par

The dataset is an essential part in promoting the development of RGBT tracking. Early works develop some small-scale datasets including OSU-CT~\cite{Davis07cviu} and LITIV~\cite{Torabi12cviu}, which contain six and nine RGBT videos respectively. 
Li et al.~\cite{Li16tip, Li17acmmm, Li2018pr} construct three larger datasets which include 50, 210 and 234 RGBT videos respectively, and also annotate some challenge attributes for challenge-based performance evaluation.
With increasing interests of RGBT tracking, the VOT challenges~\cite{Kristan2019a} take RGBT tracking as a new challenge and introduce the VOT-RGBTIR2019 dataset elaborately designed based on RGBT234.
However, several issues are remained in development and assessment of RGBT tracking algorithms.

{\flushleft \bf Lack of a large-scale dataset.} 
Existing RGBT video datasets have less than 300 sequences, as shown in Fig.~\ref{fig:bubble}. Deep learning models have dominated the research field of RGBT tracking while these small-scale datasets would limit the potentials of deep RGBT tracking models. For example, when we evaluate a tracker on RGBT234, a modern way is to initialize tracking model by pre-trained deep classification models on ImageNet dataset and then fine-tune it using GTOT dataset~\cite{Li2019iccvw}. The intrinsic differences among different tasks and the small-scale size of GTOT might result in suboptimal tracking performance~\cite{LaSOT}.

{\flushleft \bf Single imaging platform.}
RGBT sequences in each one of the existing datasets are captured by a single imaging platform. For example, for GTOT, the recording system consists of an online thermal imager and a CCD camera, which are mounted by a tripod. Due to such setup, the imaging view needs to be fixed in capturing one sequence and the scenes also need to be carefully selected to make the homography assumption effective~\cite{Li16tip}. The imaging hardware for RGBT210 consists of a turnable platform, a thermal infrared imager and a CCD camera, whose horizon has limited scenes and categories~\cite{Li2018pr}. These single imaging platforms largely restrict the diversity in data creation because of suffering from same imaging parameters and single kind of imaging setup.

{\flushleft \bf Limited number of scenes and categories.}
RGBT tracking is to track an arbitrary object in an arbitrary video sequence, and reasonable number of scenes and categories is thus of vital importance for the fair evaluation of different algorithms. Nevertheless, existing RGBT tracking datasets suffer from small number of scenes and categories. As a result, a reasonable number of scenes and categories is desired for more reliable evaluation results.

%{\flushleft \bf Short-term tracking.}
%Long-term tracking focuses on locating an object in a relative long period,  inwhich the object might disappear and re-enter the view. It is more practical and thus attracts much attention. In existing RGBT tracking datasets, however, the longest sequence length is less than 5000 frames and the object almost always appears in the video frame. The evaluation on the short-term tracking setting limit the research and deployment of RGBT tracking in practice.

\begin{table*}[t]
	\caption{Comparison of our LasHeR with recently published benchmarks.}
	\label{tb:comparision_benchmarks}
	\renewcommand\arraystretch{1.2}
	\begin{center}\scriptsize
		
		\begin{tabular}{c|cccccccc}
			\toprule
			Benchmark &Sequences &\begin{tabular}[c]{@{}l@{}}Average \\ frames\end{tabular} &\begin{tabular}[c]{@{}l@{}}Max \\ frames\end{tabular} &\begin{tabular}[c]{@{}l@{}}Total \\ frames\end{tabular} &\begin{tabular}[c]{@{}l@{}}Object \\ classes\end{tabular} &\begin{tabular}[c]{@{}l@{}}Num. of \\ attributes\end{tabular}&\begin{tabular}[c]{@{}l@{}}Multi-platform  \\ imaging\end{tabular} & \begin{tabular}[c]{@{}l@{}}Training \\ dataset\end{tabular} \\
			%\midrule
			%\midrule
			\hline
			\hline
			GTOT~\cite{Li16tip} &50 &157 &376 &7.8K &9 &7 & {\ding{55}}& {\ding{55}}\\
			RGBT210~\cite{Li17acmmm}&210 &498 &4140 &104.7K &22 &12 &{\ding{55}}& {\ding{55}} \\
			RGBT234~\cite{Li2018pr}&234 &498 &4140 &116.7K &22 &12 &\ding{55}& {\ding{55}} \\
			VOT-RGBTIR2019~\cite{Kristan2019a}&60 &334 &1335 &20.1K &13 &12 &\ding{55}& {\ding{55}}\\
			CAMEL~\cite{CAMEL}&30 &747 &2008 &22.4K &5 &- &\ding{51}& {\ding{55}}\\
			KAIST~\cite{KAIST}&41 &2702 &4708 &95K &3 &- &\ding{55}& {\ding{51}}\\
			\hline
			LasHeR &1224 &600 &12862 &734.8K &32 &19 &\ding{51}& {\ding{51}}\\
			
			\bottomrule
		\end{tabular}
	\end{center}
	%\vspace{-0.5cm}
\end{table*}

{\flushleft \bf Miss some real-world challenges.}
A good dataset should include as many real-world challenges as possible which could drive the tracker to handle them robustly in practice. However, several real-world challenges are missing in existing RGBT tracking datasets, including hyaline occlusion, frame lost, high illumination, abrupt illumination variation, out of view, similar appearance and aspect ratio change. For example, thermal cameras are usually equipped with the non-uniformity correction scheme. As a result, a fragment of thermal frames is lost and the state (e.g., appearance, location and scale) of target might change abruptly. 

To address the above issues, we provide a {\bf La}rge-{\bf s}cale {\bf H}igh-diversity b{\bf e}nchmark for {\bf R}GBT tracking (LasHeR) with the following contributions.
\begin{itemize}
\item LasHeR consists of 1224 visible and thermal infrared video pairs with more than 730K frame pairs in total. Each frame pair is spatially aligned and manually annotated with a bounding box, making the dataset well and densely annotated. It will play a significant role in both the training of deep RGBT trackers and the comprehensive evaluation of RGBT tracking methods.

\item LasHeR is highly diverse capturing from a broad range of object categories, camera viewpoints, scene complexities and environmental factors across seasons, weathers, day and night. Induced by real-world applications, several new challenges are taken into consideration in data creation. It will promote the research and development of practical tracking algorithms.

\item The unaligned version of LasHeR has also been released for the practicality of RGBT tracking. We design a scheme to automatically generate high-quality ground truths of the unaligned LasHeR by using the annotated ground truths in LasHeR and the transformations between two modalities. We hope the open of unaligned LasHeR will be beneficial to attracting the research interest for alignment-free RGBT tracking, which is a more practical task in real-world applications. 

\item Comprehensive evaluation and analysis of different RGBT tracking algorithms are conducted on LasHeR dataset. We not only make the research space in RGBT tracking more clear by the comparison of 12 RGBT tracking algorithms, but demonstrate the effectiveness of LasHeR in the training of deep RGBT trackers. 

\end{itemize}

\begin{figure*}[htbp]
	\vspace{-0.4cm}
	\centerline{\includegraphics[width=0.8\linewidth]{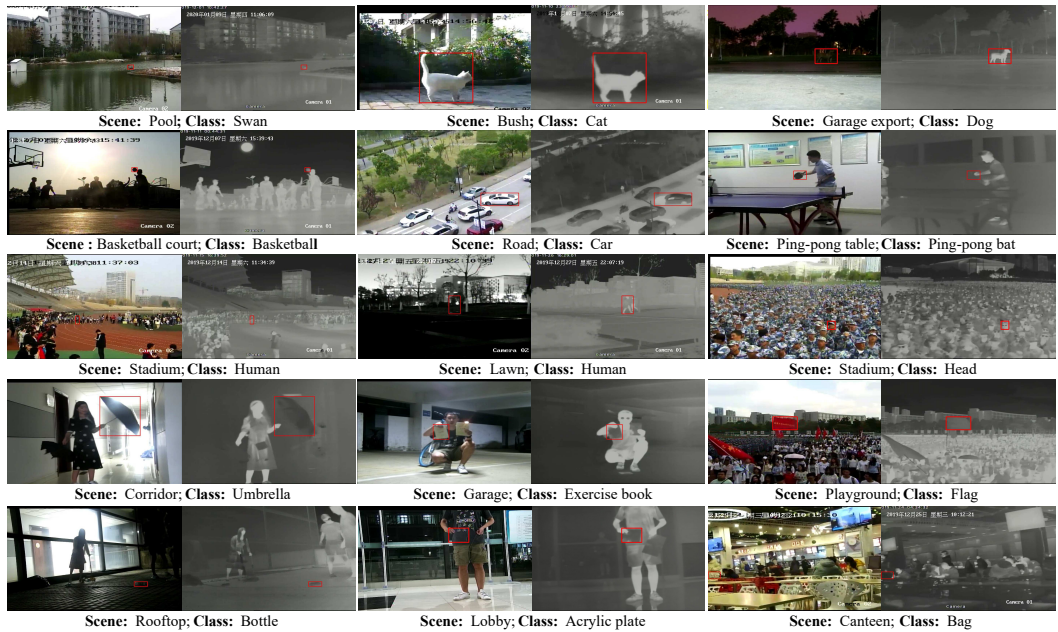}}
	\caption{Example frame pairs in LasHeR. Here we list examples of sequence images of different scenes and target classes to show the diversity of LasHeR.}
	\label{fig::scene-class}
	%\vspace{-0.3cm}
	\setlength{\belowcaptionskip}{-0.3cm}
	\vspace{-0.3cm}
\end{figure*}

\section{Related Work}
In this section, we review RGBT tracking benchmarks that are relevant to our work. At the same time, we also briefly review some of the current excellent RGBT tracking methods. For a more detailed introduction to the RGBT tracking methods, please refer to the survey~\cite{Zhang20survey}.
% for RGBT tracking methods.

\subsection{RGBT Tracking Datasets}

{\bf OSU-CT} dataset~\cite{Davis07cviu} contains six RGBT video sequence pairs recorded from two different locations with only people moving. {\bf LITIV} dataset~\cite{Torabi12cviu} contains 9 RGBT sequences which also suffers from the problems of limited size, low diversity and high bias. {\bf GTOT} dataset includes 50 RGBT video clips under different scenarios and conditions, and seven challenges are taken into account to increase the diversity.
To create a reasonable size and more challenging RGBT video dataset, Li et al.~\cite{Li17acmmm} introduce the {\bf RGBT210} dataset which contains 210 RGBT videos with highly accurate alignment and annotations of 12 challenges. 
Li et al.~\cite{Li2018pr} extend RGBT210 to a larger dataset called {\bf RGBT234}, containing 234 RGBT videos, to improve annotation quality and include more challenges.
With increasing interests of RGBT tracking, the VOT challenges~\cite{Kristan2019a} take RGBT tracking as a new challenge and introduce the {\bf VOT-RGBTIR2019} dataset elaborately designed based on RGBT234.
However, these datasets are limited by their size, diversity and bias in the training of deep RGBT tracking algorithms and comprehensive evaluation of different algorithms.
Zhang et al.~\cite{zhang2019multi} generate a large-scale synthetic RGBT
dataset generated from 8,335 videos with 1,251,981 frames in
total, but synthetic RGBT data have a great gap with real ones.

As far as we know, the above are currently commonly used datasets in the field of RGBT object tracking. We can find that several of the real RGBT datasets are small-scale in terms of the number of sequences, target categories, and challenge diversity. It is worth mentioning that there is an inclusive relationship among the three datasets of RGBT210, RGBT234 and VOT-RGBTIR2019, which means that the total number of independent sequences available for these three datasets does not exceed 234. While the effectiveness of the pseudo RGBT tracking dataset generated by the large-scale algorithm remains to be verified. In this dataset, the thermal infrared modal data are generated from the visible images by a pix2pix model~\cite{Lichao2018Synthetic}. We have to know that the original intention of introducing thermal infrared modal data in the RGBT vision field is to use the complementary information between the two modalities, so how to ensure that the generated pseudo-thermal infrared image has corresponding complementary information when the RGB data quality is poor is a difficult problem.
 
\subsection{RGBT Tracking Methods}
Due to the complementarity of thermal infrared and visible information, RGBT tracking becomes a promising research topic in the computer vision community and has attracted much attention. We can roughly divide existing RGBT tracking methods into four categories, including sparse representation-based methods, graph-based methods, correlation filter-based methods, and deep learning-based methods. We briefly review them as follows.

Sparse representation is an effective tool to suppress noises and errors, and Li et al.~\cite{Li16tip,Li16mmm,Li17tsmcs} propose collaborative sparse representation models for RGBT tracking in Bayesian filtering framework, in which modality weights are introduced to achieve adaptive fusion of different source data.
Lan et al.~\cite{Lan2018RobustCD,Lan2019OnlineNM,Lan2019LearningMF,Lan2020ModalitycorrelationawareSR} propose discriminative learning frameworks to leverage some properties of different modalities for robust RGBT tracking. 

In graph-based methods, the target in both RGB and thermal modalities is represented by a collaborative graph, in which node weights are used to suppress background interference for robust RGBT target representation. Li et al.~\cite{Li17acmmm,Li2018pr} propose a weighted sparse representation regularized graph to take global relations among graph nodes. Li et al.~\cite{Li18eccv,2021RGBTNoise} propose a cross-modal ranking method with soft consistency and noisy labels to handle the effects of modal heterogeneity and seed noises in ranking model. In addition, Li et al.~\cite{li2018spic} propose a two-stage modality-graphs regularized manifold ranking model to refine the ranking results using a two-stage way.

Correlation filter techniques are also applied to RGBT tracking due to their good performance and high efficiency. Wang et al.~\cite{Wang2018LearningSC} present a fusion tracking method based on a soft consistency correlation filter model, in which both collaboration and heterogeneity are taken into account.  Zhai et al.~\cite{Zhai2019FastRT} propose to use the low-rank constraint to learn correlation filters jointly for cross-modal fusion. Yun et al.~\cite{Yun2019DiscriminativeFC} propose a discriminative fusion correlation learning model to improve DCF-based tracking performance. Apart from the above works which purely utilize correlation filters, there are some studies to combine correlation filters with other techniques. For example, Luo et al.~\cite{Luo2019ThermalIA} propose a tracking-before-fusion framework which consists of two modules, including a correlation filter based tracking module and a histogram based tracking module. 

Deep learning is well known for being able to learn powerful feature representations from large-scale datasets~\cite{Xu2018RelativeOT,Li2018FusingTC}. One research stream is to employ Siamese networks~\cite{Zhang2019ObjectFT,Zhang2019SiamFTAR,Zhang2020DSiamMFTAR,Zhang2019Decisionlevel} which employs Siamese network for RGBT tracking. These methods have fast tracking speed, but are usually weak in representing low-resolution objects, which are common in RGBT tracking. The other main research stream is in MDNet frameworks~\cite{Zhang2018LearningMC,zhu2020quality,zhang2020object,Li2019iccvw,zhu2019acmmm,gao2019deep,Li2020ChallengeAwareRT,Wang2020CrossModalPF,Li2020DMCNet,Li2020MANet++,2021Multimodalcross,zhu2021rgbt,2020M5L}, which performs different fusion strategies to utilize complementary benefits of RGB and thermal data. Such kinds of methods receive robust tracking results but have low efficiency, and the tracking capacity is limited by MDNet which bases on VGG network. 
Zhang et al.~\cite{zhang2019multi} propose multi-fusion strategy in DiMP framework~\cite{DiMPBhat2019LearningDM} which bases on ResNet.
Regarding decision-level fusion, Tang et al.~\cite{Cong2019DecisionLevelFT} propose a SSD based RGBT tracking framework. Zhang et al.~\cite{Zhang2021JointlyMM} combine appearance tracker with motion tracker to jointly model appearance and motion cues for RGBT tracking.

\section{LasHeR Benchmark}

To resolve the contradiction between data-hungry RGBT trackers and existing small-scale datasets, we construct a large-scale video dataset, which includes a total of 1224 pairs of RGBT sequences and the total number of frames reaches to 730K, called LasHeR in this paper. The major properties of LasHeR over several existing RGBT datasets are shown in Table~\ref{tb:comparision_benchmarks}. We analyze the details as follows.

\begin{figure*}[t]
		\centerline{\includegraphics[width=0.8\linewidth]{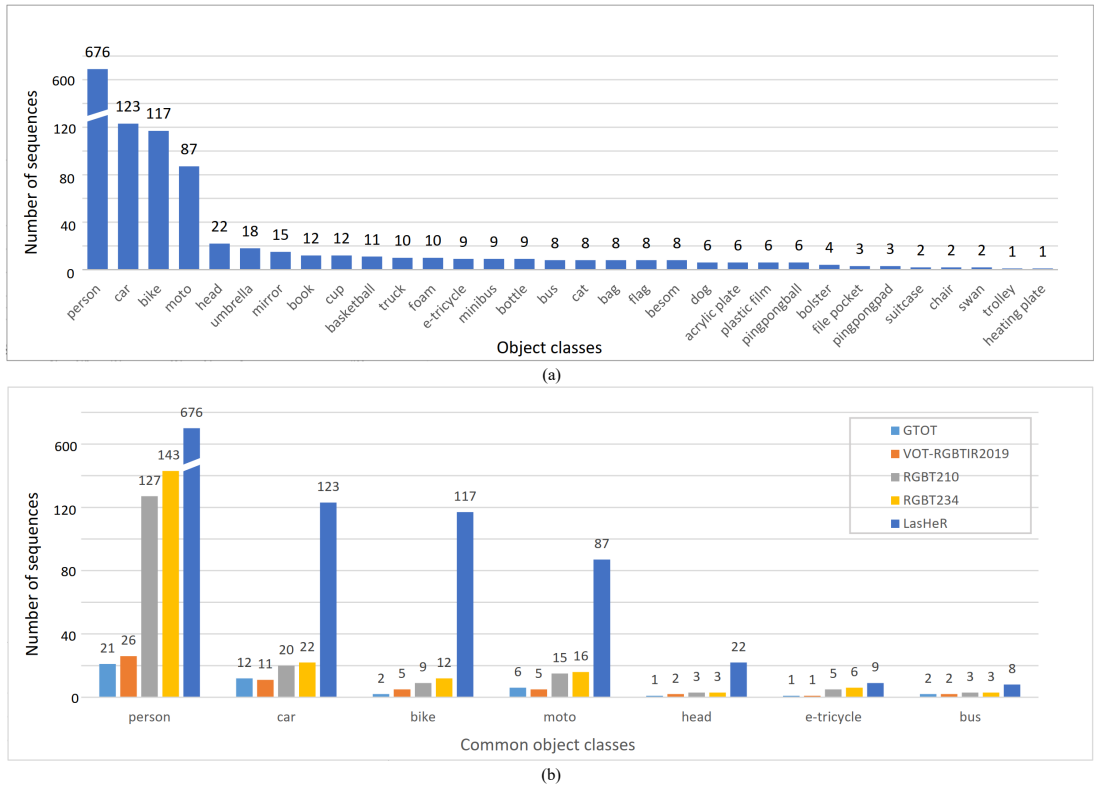}}
		\caption{Distribution of object categories on LasHeR. (a) Distribution on the entire LasHeR. (b) Distribution of common objects classes on several typical RGBT tracking datasets.}
		\label{fig:object_distribution}
		\vspace{-0.4cm}
	\end{figure*}

\subsection{Multi-platform Imaging Setup}

Existing RGBT tracking datasets base on single imaging platform, which greatly limits the diversity of data creation. For example, all frames in RGBT234~\cite{Li2018pr} are captured by a turnable platform fixed on a monitor rod. Such platform makes data have the same resolution, frame rate and limited scenes, which greatly restrict the diversity of dataset. To handle this problem, we adopt multiple types of imaging platforms. 
On one hand, we design a turnable platform like the setup in RGBT234 with a thermal image (DLS-H37DM-A) capturing thermal infrared radiation in 3-15$\mu$m band and a CCD camera (SONY EXView HAD CC) to simulate the imaging scenarios of visual surveillance and intelligent transportation. On the other hand, to capture images in a large range of scenes and conditions, we also design a hand-held platform which consists of a thermal infrared camera (DLD-J18-161) and a CCD camera (DS-2ZCN3007). It is worth noting that we can set the frame rates of the CCD camera and thermal infrared imager to the same in our system during shooting and the temporal alignment can thus be ensured. Meanwhile, the image resolution and imaging range of the two cameras are consistent, and we manually align the optical axises of two cameras to make the common horizon as large as possible, so that the image contents of these two cameras are almost same.

\par

\begin{table}[t]\scriptsize
	\caption{List and description of 19 attributes in LasHeR. Comparing with existing RGBT datasets, LasHeR introduces seven new ones indicated by italic fonts. }
	\renewcommand\arraystretch{1.1}
	\begin{tabular}{p{33 pt}p{200 pt}}
		\toprule
		\textbf{Attribute}& \textbf{Definition}\\
		\hline
		\textbf{NO}& No Occlusion - the target is not occluded.\\
		\textbf{PO}& Partial Occlusion - the target object is partially occluded.\\
		\textbf{TO}& Total Occlusion - the target object is totally occluded. \\
		\textbf{HO}& \emph{Hyaline Occlusion} - the target is occluded by hyaline object.\\
		\textbf{OV}& \emph{Out-of-View} - the target leaves the camera field of view.\\
		\textbf{LI}& Low Illumination - the illumination in the target region is low.\\
		\textbf{HI}& \emph{High Illumination} - the illumination in the target is too strong to identify the target.  \\
		\textbf{AIV}& \emph{Abrupt Illumination Variation} - the illumination of the target changes significantly. \\
		\textbf{LR}& Low Resolution - the resolution in the target region is low.\\
		\textbf{DEF}& Deformation - non-rigid object deformation.\\
		\textbf{BC}& Background Clutter - the background information which includes the target object is messy. \\
		\textbf{SA}& \emph{Similar Appearance} - there are objects of similar appearence near the target. \\
		\textbf{TC}& Thermal Crossover - the target has similar temperature with other objects or background surroundings. \\
		\textbf{MB}& Motion Blur - the target object motion results in the blur  image information.\\
		\textbf{CM}& Camera Moving - the target object is captured by moving camera.\\
		\textbf{FL}& \emph{Frame Lost} - some of thermal frames are lost.\\
		\textbf{FM}& Fast Motion - the motion of the ground truth between two adjacent frames is larger than 20 pixels.\\ 
		\textbf{SV}& Scale Variation - the ratio of the first bounding box and the current bounding box is out of the range [0.5,2].\\
		\textbf{ARC}& \emph{Aspect Ratio Change} - the ratio of bounding box aspect is outside the range [0.5,2]. \\
		\bottomrule
		
	\end{tabular}
	\label{tb::attribute}
	\vspace{-0.4cm}
\end{table}

\subsection{Multi-modal Alignment}

The videos of LasHeR proposed in this paper are collected by two different imaging platforms. Since these imaging platforms have the ability to simultaneously collect multi-modal videos with the same frame rate, the first step of data processing is to spatially align the images of the two modalities. In the task of multi-modal object detection~\cite{KAIST,MultiPedes,Weaklyalign}, multiple objects need to be localized and the global alignment should thus be required. But the problem of radial distortion would affect the alignment much especially near to image borders. Different from it, the task of visual tracking is to localize a target object, which often lies in a local region. Therefore, although our dataset still has radial distortion, we do not account for this problem and focus on the accurate alignment in a local region which covers the target object in each frame. For the two images of each frame, we estimate the homography parameters by annotating a set of matching points. Then, we keep the thermal infrared image unchanged, and apply the homography parameters on the visible one to transform it into the unified coordinate system with the thermal infrared image.

\subsection{Annotation}

A tracking dataset is desired to have high-quality densely bounding box annotations, which is essential for training a robust tracker and ensuring the fairness of performance evaluation. 
We define a deterministic annotation strategy and use the software which is called ViTBAT~\cite{ViTBAT} to annotate LasHeR where each frame is checked and manually fine-tuned to ensure the accuracy of bounding box. Given a tracking sequence for a specific tracking target, for each frame, when the target is in the view of camera, we annotate the target with a minimum bounding box covering the entire target. Otherwise, for the case where the target moves out of the view, we set the width and height of the bounding box as 0. 
Under the condition of high-accurate alignment of two modalities in LasHeR, we just need to annotate one modal image in each frame pair, since these two modalities share the same bounding box. Therefore, we annotate bounding boxes in the modality with higher quality for convenience and high-quality annotations due to the strict spatial alignment of the targets of the two modalities.

Note that some RGBT tracking datasets~\cite{Li16tip,Li2018pr} use the modality-specific annotation scheme that separately annotates RGB and thermal infrared frames with bounding boxes in each frame pair. 
This scheme is suitable for the case that the two modalities are not well aligned. The separated annotations are benefiting for the performance evaluation under a certain modality. The weakness of such annotation is that no unified ground truth labels are used to evaluate trackers and the evaluation might introduce some biases. 
Different from the above modality-specific annotation scheme, the modality-shared annotation is suitable for the case that the two modalities are well aligned. It can avoid the evaluation bias since the ground truth labels are unified, but would introduce large annotation errors if two modalities are not aligned. 
In GTOT~\cite{Li16tip} and RGBT234~\cite{Li2018pr} datasets, there are some annotation errors and thus the modality-specific annotation scheme is used, while our data are well aligned and thus adopt the scheme of modality-shared annotation.
%

%
 %\textcolor[rgb]{0,0,1}{That's to say that we just need to annotate one modal image in each frame pair, since these two modalities share the same bounding box. Therefore, we annotate bounding boxes in the modality with higher quality for convenience due to the strict spatial alignment of the targets of the two modalities.}
%

\begin{figure*}[htbp]
	\setlength{\abovecaptionskip}{-0.4cm}
	\centerline{\includegraphics[width=0.8\linewidth]{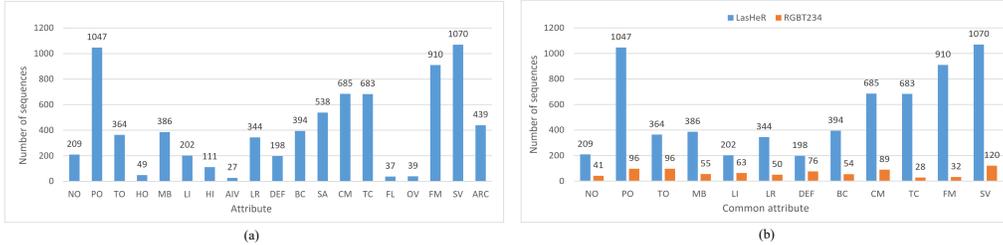}}
	\caption{Distribution of Attributes on LasHeR. (a) Distribution on the entire LasHeR. (b) Distribution of common Attributes on RGBT234 and LasHeR.}
	\label{fig:attribute_distribution}
	\vspace{-0.3cm}
\end{figure*}

\subsection{Data Statistics}

The diversity of LasHeR is embodied in the classes of objects, the types of scenes, and the imaging angles. Based on the multi-platform imaging setup, we are able to collect video pairs from a broad range of object categories, camera viewpoints, scene complexities and environmental factors across seasons, weathers, day and night. To clarify the advantages of LasHeR, we analyze the diversity from the following aspects.

{\bf Scene type}.
The complexity and type of a scene are key factors in improving the diversity of dataset. To this end, we capture videos in more than 20 scenes with different characteristics in both indoor and outdoor environments. Indoor scenes include teaching building, parking garage, corridor, canteen, etc., and outdoor scenes include basketball court, stadium, pool and road, etc. These scenes are with different complexities and thus bring some difficulties for visual trackers. Fig.~\ref{fig::scene-class} shows some example scenes in LasHeR.

{\bf Object category}. 
RGBT tracking is to locate arbitrary objects in videos, and the number of object categories should be as many as possible. Existing RGBT tracking datasets~\cite{Li16tip,Li2018pr} contain no more than 22 categories, as shown in Table~\ref{tb:comparision_benchmarks}. To improve the diversity, our LasHeR contains 32 types of target objects, including a rich variety of rigid objects and non-rigid objects.
%For rigid objects, LasHeR includes vehicle, basketball, ping-pong racket, etc. They have relatively fixed aspect ratios but often have challenges such as fast motion and motion blur.
%For non-rigid objects, including humans, cats, flags, etc., have variable appearances, length and width, and the ratio is not fixed, which brings great challenges to tracking.
Fig.~\ref{fig:object_distribution} shows the distributions of target categories on LasHeR. Here we can observe that the data distribution on object classes in LasHeR conforms to the long tail distribution, and it is a popular research topic in practical applications. As the long tail distribution of object classes in tracking field does exist in the real-world scenarios, we believe such an imbalance of data would bring meaningful challenges and encourage the design and development of more practical and extensible RGBT trackers.

{\bf Challenge}. 
In data creation, we consider more real-world challenges than existing tracking datasets~\cite{Li16tip,Li2018pr}. The labeled challenges in existing RGBT tracking datasets include no occlusion (NO), partial occlusion (PO), total occlusion (TO), low illumination (LI), low resolution (LR), deformation (DEF), background clutter (BC), motion blur (MB), thermal crossover (TC), camera moving (CM), fast motion (FM) and scale variation (SV). In addition to them, we take more factors in data creation and also label them in sequence-level for challenge-based performance evaluation of different tracking algorithms.
The newly labeled challenges include hyaline occlusion (HO), high illumination (HI), abrupt illumination variation (AIV), similar appearance (SA), aspect ratio change(ARC), out-of-view (OV) and frame lost (FL). 
The detailed definitions of all challenge factors are presented in Table~\ref{tb::attribute}. And Fig.~\ref{fig:attribute_distribution} shows the distribution of the above challenges on the LasHeR sequences and the comparison with RGBT234 in the common attributes. It is not difficult to find that the proposed LasHeR is a dataset with intensive challenges, and has a great advantage over the existing RGBT tracking dataset in terms of scale.

\begin{figure}[htbp]
	\vspace{-0.3cm}
	\centerline{\includegraphics[width=0.9\linewidth]{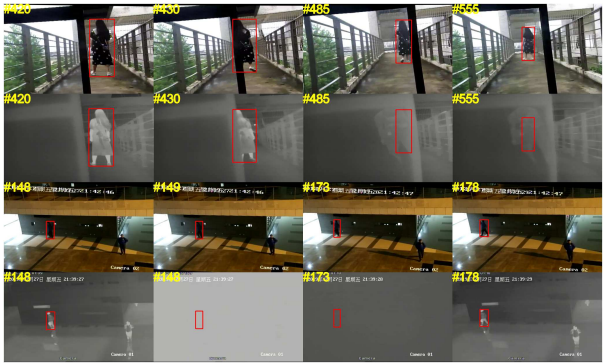}}
	\caption{Two sequence examples that show the HO and FL challenges respectively.}
	\label{fig:HOFL}
	
	\setlength{\abovecaptionskip}{-0.4cm}
	%\vspace{-0.3cm}
\end{figure}

Herein, we explain two uncommon challenges for clarity. First, it is known that thermal infrared radiation can only pass through the intervals between atoms and molecules, but cannot pass through the interior of atoms and molecules. For hyaline objects like transparent glass and acrylic, which have tight structures, thermal infrared radiation cannot pass through. Therefore, the thermal imager cannot capture the radiation of the objects behind hyaline materials. When the target object is occluded by hyaline objects, it is invisible in thermal modality but visible in visible modality. We call this challenge as hyaline occlusion.  Second, we observe that existing uncooled thermal cameras often have built-in non-uniformity correction schemes, the purpose of which is to ensure the accuracy of temperature information. Within a few seconds after the non-uniformity correction occurs, the thermal camera will stop imaging and some frames are thus lost. we call this challenge as frame lost. 
Note that, in such cases, the thermal images still exist and keep alignment with visible ones, but their image contents are unchanged or blank. Therefore, we can easily annotate the FL challenge according to the above phenomenon and do not need to provide or copy other information for thermal data. The most important thing is that during the entire processing, we always ensure that each frame pair of the two modalities is synchronized over time.  The two sequences shown in Fig.~\ref{fig:HOFL} visualize these two challenges.

 \begin{figure}[htbp]
	\centerline{\includegraphics[width=\linewidth]{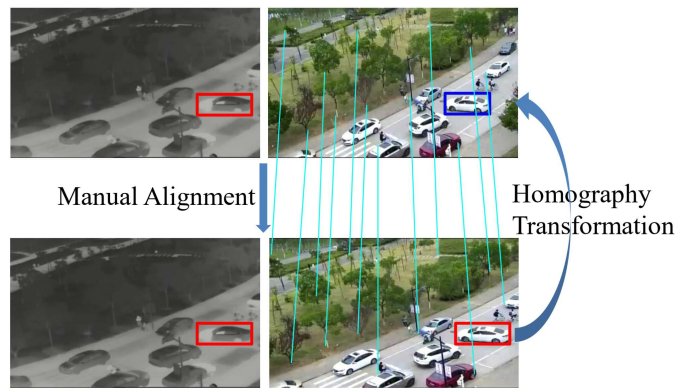}}
	\caption{Process of the generation of ground truth of unaligned RGBT pair. Herein, the same colors indicate the same ground truths.}
	\label{fig:GTalignment}
	\vspace{-0.4cm}
\end{figure}

\subsection{Unaligned LasHeR}
As we all know, the alignment of RGB and thermal images is labor- and time-consuming. It also restricts the application of RGBT tracking. Some efforts are devoted to developing RGB and thermal imaging systems with optical registrations~\cite{Gundogdu_2015_CVPR_Workshops,Zhang2019WeaklyAC,Arar2020UnsupervisedMI}, which are usually very expensive, and alignment errors are sometimes inevitably introduced~\cite{Zhang2019WeaklyAC}. Therefore, the task of alignment-free RGBT tracking is essential. We define it as follows. Given the initial state of target in unaligned RGBT videos, alignment-free RGBT tracking is to estimate states of the target in subsequent frames using both information of RGB and thermal data collaboratively. We believe that the research and development of alignment-free RGBT tracking will be beneficial to the practicality of RGBT tracking.

 To this end, we also release the unaligned version of LasHeR. Since RGB and thermal frames are unaligned, their ground truths are not shared. Note that re-annotating them is not only labor- and time-consuming but also introduces more annotation bias due to the existence of bad imaging equality. Therefore, we design a scheme to automatically generate high-quality ground truths  of  the  unaligned  LasHeR  by  using  the  annotated ground truths in LasHeR and the transformations between two modalities.

 %As we mentioned before, for the alignment of the two modal images in the target tracking task, we use linear transformation to locally align the target area. The linear transformation here includes the same translation and zooming of each pixel of the entire image. Therefore, we can obtain locally aligned RGB images from the original unaligned RGB images taken by the CCD camera. And we know that the homography matrix can meet the requirements of linear transformation. 
 %Note that after the alignment phase, we already have all of the aligned and unaligned images of this dataset but lack the annotation of unaligned RGB images of the unaligned version dataset. So the only thing that needs to be done is to obtain the accurate GT of the unaligned dataset by linearly transforming the GT of the aligned LasHeR. Due to the GTs can be regarded as the regional sub-image of the entire image, the homography matrix can be used to perform the linear transformation we expect.
 % So we thought that for the locally aligned images by linear transformation and its corresponding bounding box (which can be regarded as the regional sub-image of the entire image), the homography matrix can be used to perform the linear transformation we expect. 

Homography estimation methods can be roughly divided into two categories, including traditional methods and deep learning methods. Some of traditional methods ~\cite{Heask, Fourierlucas, Lucas1981AnII} achieve pixel-to-pixel matching by shifting or warping the images relative to each other. Other methods ~\cite{Wu2007AnIR, Rublee2011ORBAE} first extract keypoints in each image using local invariant features (e.g. SIFT), and then use feature matching to calculate the correspondence between the two sets of keypoints. Finally, they usually use the RANSAC scheme to find the best homography estimation. Deep learning methods can be categorized into the supervised ~\cite{Hu2018WeaklysupervisedCN, Hu2019DualStreamPR} and the unsupervised settings ~\cite{Le2020DeepHE, Zhang2020ContentAwareUD}. Supervised methods rely on ground truth data for the registration parameters to train the network which is extremely hard to collect. Unsupervised registration techniques often focus on the specific challenges that how to design the unsupervised loss. 
	
In our work, we aim to estimate the homography transformation by employing matches between aligned and unaligned RGB images, which can be accurately achieved by SIFT features. Therefore, we use SIFT based homography estimation to automatically generate the ground truth boxes of unaligned RGBT pairs.  Fig.~\ref{fig:GTalignment} shows the process of the generation of the ground truth (GT) of unaligned LasHeR from the annotated GT of LasHeR.

\begin{table}[t]\scriptsize
	
	\setlength{\abovecaptionskip}{-0.1cm}
	\caption{Detailed information of evaluated trackers. Representation: HoG - Histogram of Oriented Gradients, Color - Color name features, Deep - Deep models. Year: publish years.}
	\label{tab:trackers_information}
	\linespread{1.5}
	\renewcommand\arraystretch{1.2}
	\begin{center}
		
		\begin{tabular}{p{85 pt}cc cp{25 pt}<{\centering}}
			\toprule
			\multirow{2}{*}{ }
			& \multicolumn{3}{c}{Representation} & \multicolumn{1}{l}{ }\\
			\cline{2-4} 
			%\cline{5-5}
			&\multicolumn{1}{c}{{HoG}}
			& \multicolumn{1}{c}{{Color}} &\multicolumn{1}{c}{{Deep}} &\multicolumn{1}{c}{{year}}\\
			\midrule
			\midrule
			SGT~\cite{Li17acmmm} &\checkmark   &\checkmark&  & 2017\\
			CMR~\cite{Li18eccv} &\checkmark &\checkmark  &  & 2018\\
			SGT++~\cite{Li2018pr} &\checkmark &\checkmark  &  & 2018\\
			MANet~\cite{Li2019iccvw} &   & & \checkmark & 2019 \\
			DAPNet~\cite{zhu2019acmmm}&   & & \checkmark & 2019 \\
			DAFNet~\cite{gao2019deep} &  & & \checkmark & 2019 \\
			mfDiMP~\cite{zhang2019multi} &  &  & \checkmark & 2019 \\
			MaCNet~\cite{zhang2020object} &  &  & \checkmark & 2020 \\
			CAT~\cite{Li2020ChallengeAwareRT} &  &  & \checkmark & 2020 \\
			FANet~\cite{zhu2020quality} &   & & \checkmark & 2020 \\
			
			DMCNet~\cite{Li2020DMCNet} &  &  & \checkmark & 2020 \\
			MANet++~\cite{Li2020MANet++} &  &  & \checkmark & 2021 \\

			\bottomrule
		\end{tabular}
	\end{center}
	\vspace{-0.5cm}
\end{table}

 %Note that we adjust RGB image with thermal image fixed in manual annotation, and the GTs of unaligned and aligned thermal images are the same. Then, we compute the homography transformation using the matching points between unaligned and aligned RGB images. Finally, we transform the GT of RGB image according to the homography transformation to the GT of unaligned RGB image. Since the content of two RGB images is almost the same, the homography transformation is very accurate, which guarantees the high-quality generation of GT. 
 
The acquisition mechanisms of the two kinds of imaging platforms are similar, and we describe the overall process of data creation as follows. First, we capture the unaligned RGBT data through the designed imaging platforms. Second, we achieve accurate local alignment around targets using the homograpy transformation technique to generate the aligned LasHeR dataset. Finally, we generate ground truth boxes of unaligned RGBT pairs from aligned ones which have been annotated accurately using the homograpy transformation technique. In this way, we avoid the second annotations to relieve manual labor while obtaining high-quality annotations for unaligned RGBT pairs. Since the content of these two RGB images is almost the same, the homography transformation is very accurate, which guarantees the high-quality generation of ground truth.

%\textcolor[rgb]{0,0,1}{In terms of the contents of the two modalities in the unaligned version, the scenario that an object is present in one modality but not the other might occur but it would not affect tracking algorithms much and the reason is as follows. First, we set the imaging parameters of two cameras to the same, such as image resolution and imaging range. Second, we manually align the optical axises of two cameras to make the common horizon as large as possible. The image contents of these two cameras are almost same due to the used above two strategies. }

\begin{figure*}[htbp]
	\setlength{\abovecaptionskip}{-0.3cm}
	\centerline{
	\includegraphics[width=\linewidth]{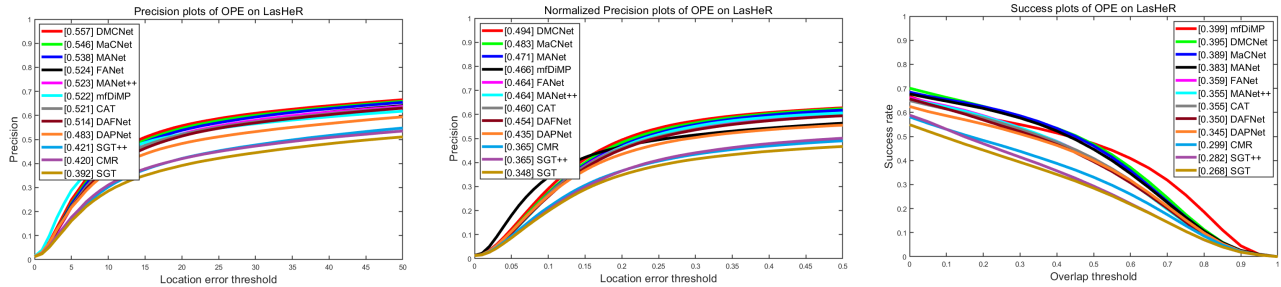}}
	\caption{Evaluation result on entire LasHeR using precision, normalized precision and success plots, where the representative scores are presented in the legend.}
	\label{fig:LasHeR_evaluation}
	\vspace{-0.3cm}
\end{figure*}

\subsection{Evaluated Trackers}

We evaluate 12 RGBT tracking algorithms on LasHeR to provide a comprehensive platform of performance analysis. Deep RGBT trackers include MANet~\cite{Li2019iccvw}, DAPNet~\cite{zhu2019acmmm}, MaCNet~\cite{zhang2020object}, DAFNet~\cite{gao2019deep}, FANet~\cite{zhu2020quality},  DMCNet~\cite{Li2020DMCNet}, MANet++~\cite{Li2020MANet++} and mfDiMP~\cite{zhang2019multi}. RGBT trackers based on handcrafted features include SGT~\cite{Li17acmmm}, CMR~\cite{Li18eccv} and SGT++~\cite{Li2018pr}. We present the details of these RGBT trackers in Table~\ref{tab:trackers_information}.

\subsection{Evaluation Metrics}

Precision rate and success rate are two widely used metrics to evaluate tracking algorithms. In our benchmark, we utilize them for tracking evaluation. %In order to reduce the impact of target scale on precesion metric, we also use normalized precesion to evaluate trackers.

\begin{itemize}
	\item {\bf Precision rate (PR)}. The precision rate is to calculate the percentage of frames where the distance between the predicted position and the ground truth is within a certain threshold range. In this work, we set the threshold to 20 pixels to compute the representative PR score.

\item {\bf Normalized precision rate (NPR)}. Since the precision metric is easily affected by the image resolution and the size of the bounding box, we further normalized the precision as in ~\cite{Mller2018TrackingNetAL} as the second metric. For detailed calculation of NPR, please refer to ~\cite{Mller2018TrackingNetAL}.

	\item {\bf Success rate (SR)}.  The success rate is to calculate the ratio of successful frames where the overlap between the predicted bounding box and the ground truth is greater than a certain threshold. In this work, we employ the area under curve to compute the representative SR score.
\end{itemize}

\section{Evaluation and Analysis}

In this section, we adopt two evaluation protocols for the evaluation of tracking algorithms. First, we use the entire dataset as a testing set, and evaluate 12 RGBT tracking methods on the entire LasHeR. Then, in order to provide a large-scale set for the training of deep trackers, we split LasHeR into a training subset and a testing subset, and the evaluation of trackers is performed on the testing subset. Finally, the retrianing experiments are conducted using the training subset of LasHeR.

\subsection{Overall Evaluation Results}
We first present the evaluation results on the entire LasHeR dataset, where all 1224 sequence pairs are used as testing set for large-scale evaluations. 
We summarize the overall evaluation results of these trackers with the precision, normalized precision and success plots, as shown in Fig.~\ref{fig:LasHeR_evaluation}. 
From the results, we can see that DMCNet achieves the best precision score of 0.557 and normalized precision score of 0.494, while mfDiMP combines the IoU-Net based architecture from ATOM~\cite{ATOM}, achieving the top success score of 0.399. DMCNet based on MDNet~\cite{MDNetNam2016LearningMC} is trained in an online fashion, which leads to a slow tracking speed due to high computational complexity, although the tracker achieves the top precision. 
While based on DiMP~\cite{DiMPBhat2019LearningDM}, mfDiMP learns the classification model in an off-line manner, which can achieve a fast tracking speed during testing.

Compared with the existing RGBT tracking datasets, the overall performance of these trackers on the proposed large-scale LasHeR has a significant degradation. For example, DMCNet achieves the PR score of 0.839 and the SR score of 0.593 on RGBT234, while the PR and SR scores of mfDiMP on RGBT210 are respectively up to 0.785 and 0.559. Such a significant performance drop further proves the difficulty of the proposed dataset, which is inseparable from the diversity of challenges and scene categories of the dataset. As mentioned before, one of the most notable features of LasHeR is that the tracking data captured by mobile platforms are closer to the real-world tracking scenarios. 
The regression of these trackers on LasHeR also indicates that trackers trained on existing datasets are not enough to cope with the challenges of real-world tracking applications due to limited diversity in challenges and scenes of existing datasets. In view of the above-mentioned problems, a high-diversity and challenging dataset is supposed to be proposed to provide great space for the research and development of RGBT trackers.

\subsection{Challenge-based Evaluation Results}

To specifically analyze the advantages of trackers from the perspective of different challenges on LasHeR, we evaluate all 12 trackers under 19 challenge attributes. We present the attribute-based PR and SR scores on all 19 challenge attributes in Table~\ref{tab:challenge-based}.

 \begin{table*}[htbp]\scriptsize
	\caption{Challenge-based precision and success scores of 12 trackers on LasHeR, including MANet~\cite{Li2019iccvw}, DAPNet~\cite{zhu2019acmmm}, MaCNet~\cite{zhang2020object}, DAFNet~\cite{gao2019deep}, FANet~\cite{zhu2020quality}, CAT~\cite{Li2020ChallengeAwareRT}, MANet++~\cite{Li2020MANet++}, DMCNet~\cite{Li2020DMCNet}, mfDiMP~\cite{zhang2019multi}, SGT~\cite{Li17acmmm}, CMR~\cite{Li18eccv} and SGT++~\cite{Li2018pr}. The last row shows the speed of these trackers. The \textcolor[rgb]{1,0,0}{red}, \textcolor[rgb]{0,0,1}{blue} and \textcolor[rgb]{0,1,0}{green} fonts represent the top three values respectively.}
	\label{tab:challenge-based}
	\renewcommand\arraystretch{1.1}
	\begin{center}
		\setlength{\tabcolsep}{4.7mm}{
			\begin{tabular}{c ccccccccc ccc}
				\toprule
				&MANet &DAPNet &MaCNet &DAFNet &FANet &CAT \\
				\midrule
				NO &0.674/0.472&0.621/0.439 &\textcolor[rgb]{0,0,1}{0.692}/\textcolor[rgb]{0,0,1}{0.492}&0.646/0.443 &0.647/0.446 &0.656/0.454 \\
				PO &\textcolor[rgb]{0,1,0}{0.509}/\textcolor[rgb]{0,1,0}{0.369} &0.454/0.329 &\textcolor[rgb]{0,0,1}{0.517}/\textcolor[rgb]{0,0,1}{0.373}&0.486/0.334 &0.496/0.342 &0.491/0.339 \\
				TO &\textcolor[rgb]{0,1,0}{0.410}/\textcolor[rgb]{0,1,0}{0.294} &0.350/0.253 &\textcolor[rgb]{1,0,0}{0.428}/\textcolor[rgb]{0,0,1}{0.307} &0.384/0.265 &0.384/0.268 &0.385/0.265 \\
				HO &\textcolor[rgb]{0,0,1}{0.312}/\textcolor[rgb]{0,0,1}{0.314} &0.234/0.261&\textcolor[rgb]{1,0,0}{0.321}/\textcolor[rgb]{1,0,0}{0.328} &0.210/0.228 &0.223/0.254 &0.239/0.250 \\
				OV &0.406/0.281 &\textcolor[rgb]{0,1,0}{0.415}/0.274 &\textcolor[rgb]{0,0,1}{0.453}/\textcolor[rgb]{1,0,0}{0.319} &0.386/0.238 &0.357/0.230 &0.340/0.203 \\
				LI &\textcolor[rgb]{0,1,0}{0.450}/\textcolor[rgb]{0,0,1}{0.333} &0.408/0.302 &\textcolor[rgb]{0,0,1}{0.452}/\textcolor[rgb]{0,1,0}{0.329} &0.432/0.300 &0.443/0.308 &0.425/0.297\\
				HI &\textcolor[rgb]{0,1,0}{0.589}/\textcolor[rgb]{0,0,1}{0.414} &0.530/0.373 &\textcolor[rgb]{0,0,1}{0.591}/\textcolor[rgb]{0,0,1}{0.414} &0.508/0.336 &0.558/0.379 &0.543/0.368\\ 
				AIV &\textcolor[rgb]{0,0,1}{0.454}/\textcolor[rgb]{0,0,1}{0.345} &0.395/0.288 &0.440/\textcolor[rgb]{0,1,0}{0.328} &0.433/0.291 &\textcolor[rgb]{0,1,0}{0.451}/0.320 &0.449/0.310\\ 
				LR &0.532/\textcolor[rgb]{0,1,0}{0.331} &0.453/0.285 &\textcolor[rgb]{0,1,0}{0.532}/\textcolor[rgb]{0,0,1}{0.333} &0.512/0.303 &\textcolor[rgb]{0,0,1}{0.534}/0.320 &0.513/0.305 \\
				DEF &0.425/0.346 &0.418/0.330 &\textcolor[rgb]{0,0,1}{0.448}/\textcolor[rgb]{0,0,1}{0.362} &0.435/0.327 &0.403/0.314 &0.420/0.323\\ 
				BC &0.454/\textcolor[rgb]{0,1,0}{0.342} &0.416/0.308 &\textcolor[rgb]{0,1,0}{0.463}/\textcolor[rgb]{0,0,1}{0.344} &0.439/0.313 &\textcolor[rgb]{0,0,1}{0.466}/0.329 &0.443/0.317\\ 
				SA &0.479/\textcolor[rgb]{0,1,0}{0.345} &0.436/0.313 &\textcolor[rgb]{0,0,1}{0.495}/\textcolor[rgb]{0,0,1}{0.356} &0.478/0.324 &0.478/0.330 &0.469/0.322 \\
				TC &\textcolor[rgb]{0,1,0}{0.468}/\textcolor[rgb]{0,1,0}{0.332} &0.417/0.297 &\textcolor[rgb]{0,0,1}{0.479}/\textcolor[rgb]{0,0,1}{0.340} &0.448/0.303 &0.454/0.311 &0.450/0.307 \\
				MB &\textcolor[rgb]{0,1,0}{0.504}/\textcolor[rgb]{0,1,0}{0.347} &0.455/0.313 &\textcolor[rgb]{0,0,1}{0.521}/\textcolor[rgb]{1,0,0}{0.362} &0.482/0.311 &0.488/0.319 &0.489/0.317 \\
				CM &\textcolor[rgb]{0,1,0}{0.508}/0.369 &0.450/0.324 &\textcolor[rgb]{0,0,1}{0.526}/\textcolor[rgb]{0,0,1}{0.379} &0.491/0.341 &0.495/0.345 &0.496/0.343 \\
				FL &\textcolor[rgb]{0,0,1}{0.371}/\textcolor[rgb]{0,0,1}{0.252} &0.325/0.222 &0.332/0.224 &0.333/0.216 &0.319/0.189 &\textcolor[rgb]{0,1,0}{0.351}/0.203\\ 
				FM &0.489/0.362 &0.429/0.317 &\textcolor[rgb]{0,0,1}{0.498}/\textcolor[rgb]{0,1,0}{0.366} &0.466/0.325 &0.472/0.334 &0.469/0.331 \\
				SV &\textcolor[rgb]{0,1,0}{0.534}/0.379 &0.475/0.338 &\textcolor[rgb]{0,0,1}{0.541}/\textcolor[rgb]{0,1,0}{0.386} &0.504/0.339 &0.514/0.350 &0.509/0.347 \\
				ARC &0.400/0.304 &0.363/0.278 &\textcolor[rgb]{0,1,0}{0.410}/\textcolor[rgb]{0,1,0}{0.313} &0.379/0.270 &0.379/0.275 &0.390/0.280 \\
				\hline
				ALL &\textcolor[rgb]{0,1,0}{0.538}/0.383 &0.483/0.345 &\textcolor[rgb]{0,0,1}{0.546}/\textcolor[rgb]{0,1,0}{0.392} &0.514/0.350 &0.524/0.359 &0.521/0.355 \\
				\hline
				Speed [fps] &0.904 &1.475 &0.891 &\textcolor[rgb]{0,0,1}{22.208} &\textcolor[rgb]{0,1,0}{17.476} &8.938\\
				\midrule
				\midrule
				&MANet++ &DMCNet &mfDiMP &SGT &CMR &SGT++\\
				\midrule
				NO &0.653/0.449 &\textcolor[rgb]{0,1,0}{0.681}/\textcolor[rgb]{0,1,0}{0.475} &\textcolor[rgb]{1,0,0}{0.738}/\textcolor[rgb]{1,0,0}{0.572} &0.499/0.331 &0.574/0.401&0.557/0.358 \\
				PO &0.498/0.343&\textcolor[rgb]{1,0,0}{0.527}/\textcolor[rgb]{1,0,0}{0.380} &0.474/0.362 &0.392/0.272 &0.407/0.292 &0.416/0.283\\
				TO &0.382/0.265 &\textcolor[rgb]{0,0,1}{0.426}/\textcolor[rgb]{1,0,0}{0.309} &0.341/0.261 &0.302/0.212 &0.313/0.223 &0.320/0.220\\
				HO &0.252/0.253 &\textcolor[rgb]{0,1,0}{0.296}/\textcolor[rgb]{0,1,0}{0.298} &0.202/0.247 &0.217/0.227 &0.210/0.220 &0.229/0.247\\
				OV &0.392/0.235 &\textcolor[rgb]{1,0,0}{0.469}/\textcolor[rgb]{0,0,1}{0.317} &\textcolor[rgb]{0,0,1}{0.453}/\textcolor[rgb]{0,1,0}{0.316} &0.371/0.246 &0.385/0.254 &0.357/0.235\\
				LI &0.441/0.303 &\textcolor[rgb]{1,0,0}{0.481}/\textcolor[rgb]{1,0,0}{0.348} &0.408/0.310&0.344/0.250 &0.344/0.252 &0.369/0.263\\
				HI &0.568/0.379 &\textcolor[rgb]{1,0,0}{0.615}/\textcolor[rgb]{1,0,0}{0.428} &0.548/\textcolor[rgb]{0,1,0}{0.408} &0.420/0.275 &0.477/0.328 &0.465/0.290\\
				AIV &0.411/0.291 &\textcolor[rgb]{1,0,0}{0.476}/\textcolor[rgb]{1,0,0}{0.370} &0.355/0.293 &0.310/0.218 &0.312/0.242 &0.317/0.219\\
				LR &0.527/0.316 &\textcolor[rgb]{1,0,0}{0.560}/\textcolor[rgb]{1,0,0}{0.353} &0.472/0.313 &0.398/0.240 &0.428/0.267 &0.448/0.262 \\
				DEF &0.432/0.328 &\textcolor[rgb]{1,0,0}{0.473}/\textcolor[rgb]{1,0,0}{0.373} &\textcolor[rgb]{0,1,0}{0.439}/\textcolor[rgb]{0,1,0}{0.358} &0.352/0.282 &0.358/0.278 &0.360/0.284\\
				BC &0.456/0.322 &\textcolor[rgb]{1,0,0}{0.467}/\textcolor[rgb]{1,0,0}{0.349} &0.403/0.313 &0.359/0.259 &0.349/0.260 &0.364/0.250\\
				SA &\textcolor[rgb]{0,1,0}{0.482}/0.328 &\textcolor[rgb]{1,0,0}{0.511}/\textcolor[rgb]{1,0,0}{0.367} &0.445/0.339 &0.400/0.281 &0.422/0.302 &432/0.294\\
				TC &0.455/0.309 &\textcolor[rgb]{1,0,0}{0.492}/\textcolor[rgb]{1,0,0}{0.347} &0.436/0.328 &0.364/0.251 &0.385/0.271 &0.393/0.264\\
				MB &0.483/0.315 &\textcolor[rgb]{1,0,0}{0.525}/\textcolor[rgb]{1,0,0}{0.362} &0.476/\textcolor[rgb]{0,0,1}{0.351} &0.367/0.250 &0.395/0.272 &0.422/0.272\\
				CM &0.493/0.341 &\textcolor[rgb]{1,0,0}{0.541}/\textcolor[rgb]{1,0,0}{0.390} &0.501/\textcolor[rgb]{0,1,0}{0.378} &0.381/0.263 &0.400/0.288 &0.411/0.277\\
				FL &0.344/0.202 &\textcolor[rgb]{1,0,0}{0.393}/\textcolor[rgb]{1,0,0}{0.283} &0.292/\textcolor[rgb]{0,1,0}{0.226} &0.327/0.214 &0.285/0.196 &0.329/0.195\\
				FM &0.471/0.331 &\textcolor[rgb]{1,0,0}{0.513}/\textcolor[rgb]{0,0,1}{0.376} &\textcolor[rgb]{0,1,0}{0.491}/\textcolor[rgb]{1,0,0}{0.385} &0.363/0.257 &0.380/0.281 &0.388/0.270\\
				SV &0.516/0.351 &\textcolor[rgb]{1,0,0}{0.549}/\textcolor[rgb]{0,0,1}{0.389} &0.525/\textcolor[rgb]{1,0,0}{0.403} &0.401/0.269 &0.430/0.304 &0.431/0.283\\
				ARC &0.402/0.286 &\textcolor[rgb]{0,0,1}{0.432}/\textcolor[rgb]{0,0,1}{0.325} &\textcolor[rgb]{1,0,0}{0.433}/\textcolor[rgb]{1,0,0}{0.347} &0.301/0.228 &0.307/0.235 &0.316/0.237\\
				\hline
				ALL &0.523/0.355 &\textcolor[rgb]{1,0,0}{0.557}/\textcolor[rgb]{0,0,1}{0.395} &0.522/\textcolor[rgb]{1,0,0}{0.399} &0.392/0.268 &0.420/0.299 &0.427/0.282 \\
				\hline
				Speed [fps] &16.153 &2.244 &\textcolor[rgb]{1,0,0}{38.242} &0.957 &2.811 &1.192\\
				
				\bottomrule
		\end{tabular}}
	\end{center}
	\vspace{-0.3cm}
\end{table*}

 As we can observed from Table~\ref{tab:challenge-based}, when the challenges such as total occlusion, hyaline occlusion, and out-of-view  exist in the sequence, trackers are more likely to lose target causing low PR and SR scores. When target disappears, trackers are hard to find the target, partly because the target is out of searching window when it is back to view again. 
 To address these  difficult but common challenges in RGBT tracking, a potential method is to use the trajectory inference to solve the problem of target disappearance.
 In addition, when target appears again, trackers are expected to be able to recognize the target again. It requires the models to have more robust modeling of target appearance and long memory for target features.

 The challenging factors like deformation and aspect ratio changes also easily degrade tracking performance. It is because that trackers can not effectively model target appearance when it changes drastically. An effective way to handle these challenges is to update the model in tracking process. 
 When the challenges of similar appearance and thermal crossover happen, trackers are also easily failed due to the little  discriminative features to distinguish similar objects. We can alleviate this problem by learning more contextual information in appearance modeling.

For the seven newly introduced challenges in the proposed LasHeR, we can observe that trackers consistently show low tracking performance on these challenges, as shown in Table~\ref{tab:challenge-based}. Trackers perform almost worst under the challenges of healine occlusion, frame lost and aspect ratio change. The results indicate that we should put an emphasis on the research of the newly introduced challenges in RGBT tracking tasks. It will be beneficial to improving the practicality of RGBT trackers in the future.

\subsection{Qualitative Evaluation}

\begin{figure*}[htbp]
	\centerline{\includegraphics[width=0.8\linewidth]{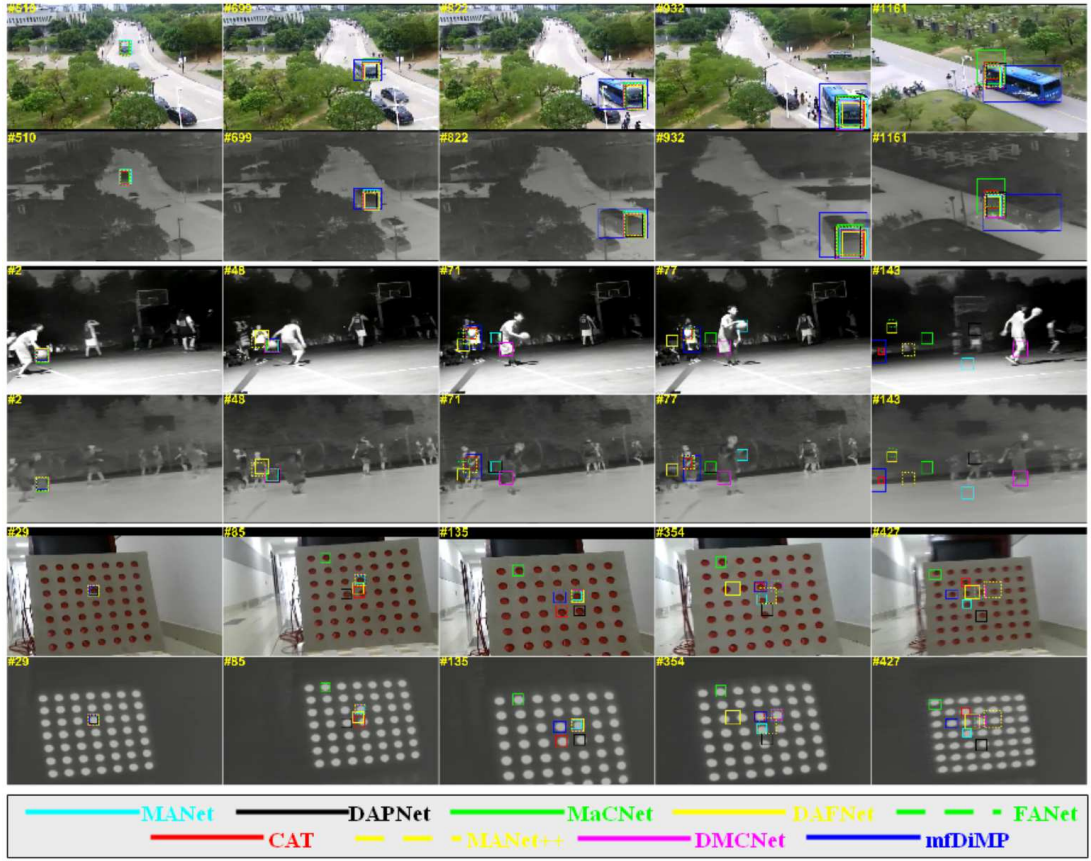}}
	\caption{Qualitative evaluation in five representative sequences: \emph{bluebuscoming}, \emph{leftbasketball}, \emph{dotat43}. The color of bounding box denotes a specific tracker. And every two rows belong to the same sequence, the upper row is from RGB images, and the images at lower row are in infrared modality.}
	\label{fig:QualitativeEva}
	%\vspace{-0.1cm}
\end{figure*}

In this part, we conduct a qualitative evaluation of nine deep trackers including MANet~\cite{Li2019iccvw}, DAPNet~\cite{zhu2019acmmm}, MaCNet~\cite{zhang2020object}, DAFNet~\cite{gao2019deep}, FANet~\cite{zhu2020quality},  DMCNet~\cite{Li2020DMCNet}, MANet++~\cite{Li2020MANet++} and mfDiMP~\cite{zhang2019multi}. We select three representative sequences in LasHeR where there are frequent challenges such as total occlusion, thermal crossover, deformation, similar appearance, and visualize the tracking results of these trackers on these sequences, as shown in Fig.~\ref{fig:QualitativeEva}.

From Fig.~\ref{fig:QualitativeEva}, we can observe that most trackers cannot cope with the challenges of aspect ratio changes and scale variation. Among them, mfDiMP handles the challenge of scale variation best, which can also be verified in the challenge-based results in Table~\ref{tab:challenge-based}. The excellent performance of mfDiMP in target scale handling benefits from the DiMP~\cite{DiMPBhat2019LearningDM}, which solves the problem of update the model of the trackers based on the Siamese framework. However, when the challenges like fast motion, similar appearance and occlusion  occur, almost all trackers tend to lose the target. In real scenes, challenges such as similar appearance and fast motion often appear at the same time. In these cases, most trackers are prone to lose their targets. To cope with the challenge of fast motion, we can think of increasing the search range, but at the same time more background interference information is introduced, causing the challenge of similar appearance more serious. Therefore, how to weigh the solutions between different challenges is a topic worthing to be studied.

\subsection{Retraining Experiment on LasHeR}

We also suggest the second evaluation protocol to evaluate RGBT tracking algorithms, under which we split LasHeR into \emph{training} and \emph{testing} subsets according to the target class distribution. The sequences in \emph{training} set are suggested to be used to train trackers and we can then assess the trackers on the \emph{testing} set. 

\begin{figure*}[htbp]
	\vspace{-0.3cm}
	\setlength{\abovecaptionskip}{-0.3cm}
	\centerline{\includegraphics[width=\linewidth]{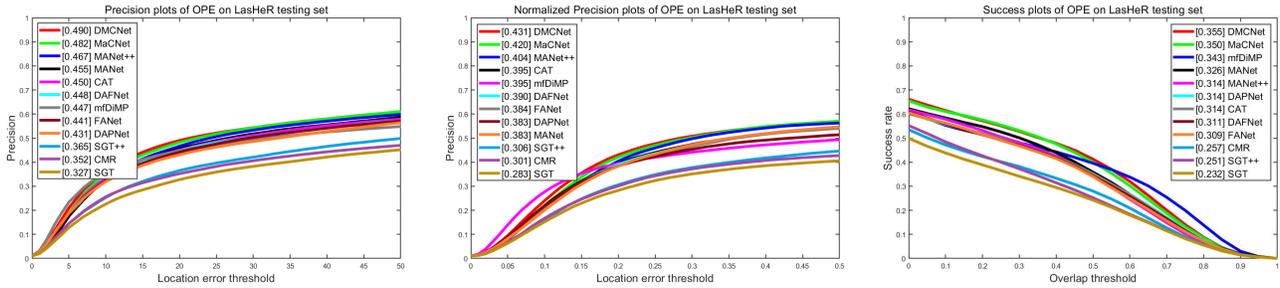}}
	\caption{Evaluation results on LasHeR \emph{testing} set using precision, normalizes precision and success plots, where the representative scores are in the legend.}
	\label{fig:LasHeRTestingSet_evaluation}
	\vspace{-0.3cm}
\end{figure*}

We evaluate 12 RGBT trackers on the \emph{testing} set to provide baselines and comparison. Same as before, each of the 12 trackers is evaluated as it is for testing without any modification or re-training. The performance is reported in Fig.~\ref{fig:LasHeRTestingSet_evaluation} with precision plot, normalized precision and success plot. Consistent results as they are in the first protocol can be observed. On the LasHeR \emph{testing} subset, the performance of all trackers has dropped significantly compared to existing datasets. DMCNet achieves the top performance with precision score as 0.490, normalized precision score as 0.431 and success score as 0.355.

In addition to the evaluation of the performance of RGBT trackers on \emph{testing} set, we conduct the retraining experiment by retraining MANet and mfDiMP on \emph{training} set to demonstrate how deep RGBT trackers can be improved using a large-scale training set. The tracker MANet designs three kinds of adapters within the network based on MDNet to jointly perform modality-shared, modality-specific and instance-aware feature learning. While the deep RGBT tracker mfDiMP using the RGB tracker DiMP as baseline tracker with offline training, uses the Siamese structure to extract image features. 

Table~\ref{tab:retraining} reports the results of MANet and mfDiMP on two  \emph{testing} sets and comparison of original trackers trained on GTOT and a larger synthetic RGBT tracking dataset respectively and the mfDiMP tracker trained on GTOT with these two deep trackers trained on the proposed LasHeR \emph{training} set. The consistent performance gains prove that a large-scale training set for deep trackers is of vital importance. It is worth noting that the retrained mfDiMP has a great improvement on LasHeR \emph{testing} set than on RGBT234, and the final PR and SR scores both exceed the best tracker DMCNet without retraining. In the comparison with the tracker trained on the large-scale synthetic dataset, it can be found that the improvement of PR and SR scores proves the necessity of a large-scale real but not synthetic RGBT tracking dataset.

Although the large number of synthetic RGBT tracking sequences used for training mfDiMP is nearly 9 times that of LasHeR \emph{training} set, the performance of the tracker has not improved but declined. It suggests that the real RGBT tracking dataset can improve RGBT trackers better. In addition, in the comparison with the performance of mfDiMP trained on GTOT, we can have the following conclusion that a large-scale dataset can be easier to train the tracker to achieve a better performance, which further proves the necessity of the proposed large-scale dataset.

\par
%As a typical MDNet-based RGBT tracking method, MANet can show good tracking performance under the training of a small-scale dataset. So we also conducted the same retraining experiment with MANet. Regrettably, the tracker MANet trained with the proposed large-scale dataset LasHeR \emph{training} set has no performance improvement on RGBT234, and although there is an improvement on the LasHeR \emph{testing} set, it is very small. We analyze the reason for this. Unlike mfDiMP, which is based on siamese feature extraction frameworks, MANet is improved based on MDNet, and the network parameters scale is relatively small, which shows that the requirements for the amount of data are relatively low, and a large amount of training data may bring negatives leading to overfitting. And we know that the proposed LasHeR  and RGBT234 have a big difference in the scene and target motion distribution. Overfitting the LasHeR training set will cause the tracker's performance on RGBT234 to decrease.
\begin{table*}[htbp]\scriptsize
	\setlength{\abovecaptionskip}{-0.3cm}
	\caption{Retraining of MANet~\cite{Li2019iccvw} and mfDiMP~\cite{zhang2019multi} on LasHeR.}
	\label{tab:retraining}
	\renewcommand\arraystretch{1.1}
	\begin{center}
		\setlength{\tabcolsep}{3mm}{
			\begin{tabular}{cc cc ccc}
				\toprule
				\multicolumn{2}{c}{ }
				& \multicolumn{2}{c}{MANet~\cite{Li2019iccvw}} & \multicolumn{3}{c}{mfDiMP~\cite{zhang2019multi} }\\
				\cline{3-7} 
				%\cline{5-5}
				\multicolumn{2}{c}{\diagbox{Testing data}{Training data}}
				& \multicolumn{1}{c}{GTOT~\cite{Li16tip}} &\multicolumn{1}{c}{\begin{tabular}[c]{@{}l@{}}LasHeR \\ \emph{training} set\end{tabular}} 
				&\multicolumn{1}{c}{\begin{tabular}[c]{@{}l@{}}Synthetic \\ dataset\end{tabular}~\cite{zhang2019multi}}
				&\multicolumn{1}{c}{GTOT~\cite{Li16tip}} &\multicolumn{1}{c}{\begin{tabular}[c]{@{}l@{}}LasHeR \\ \emph{training} set\end{tabular}}\\
				\midrule
				%\midrule
				\multirow{2}{*}{RGBT234~\cite{Li2018pr}} & Precision  &0.777 &0.810  & 0.785 &0.787 &0.842\\
				&Success &0.539 &0.569 &0.559 &0.560 &0.591\\
				\midrule
				\multirow{2}{*}{\begin{tabular}[c]{@{}l@{}}LasHeR \\ \emph{testing} set\end{tabular}} & Precision  &0.455 &0.508  &0.447  &0.474 &0.599 \\
				&Success &0.326 &0.369 &0.343 &0.364 &0.467\\

				\bottomrule
		\end{tabular}}
	\end{center}
	\vspace{-0.3cm}
\end{table*}

\begin{table*}[htbp]\scriptsize
	\setlength{\abovecaptionskip}{-0.1cm}
	\caption{Evaluation results of TransT~\cite{TransT}, SiamFC++~\cite{SiamFC++}, PrDiMP~\cite{PrDiMP}, DiMP~\cite{DiMPBhat2019LearningDM} and ATOM~\cite{ATOM} on LasHeR, LaSOT~\cite{LaSOT} and TrackingNet~\cite{Mller2018TrackingNetAL}.}
	\label{tab:RGB_results_lasher}
	\renewcommand\arraystretch{1.3}
	\begin{center}
		\setlength{\tabcolsep}{3mm}{
			\begin{tabular}{|c| c |ccc |ccc |ccc|}
				\hline
				
				\multirow{2}{*}{Methods}
				& \multirow{2}{*}{Source} & \multicolumn{3}{c|}{LasHeR } & \multicolumn{3}{c|}{LaSOT~\cite{LaSOT}} & \multicolumn{3}{c|}{TrackingNet~\cite{Mller2018TrackingNetAL}}\\
				
				\cline{3-11} 
				%\cline{5-5}
				&&PR &NPR&SR &PR &NPR &SR
				&PR &NPR &SR\\

				\hline
				%\midrule
				TransT~\cite{TransT} &CVPR2021&0.592 &0.548 &0.442 &0.690 &0.738 &0.649
				&0.803 &0.867 &0.814\\
				
				%\hline
				
				SiamFC++~\cite{SiamFC++} &AAAI2020&0.415 &0.374 &0.320 &0.547 &0.623 &0.544
				&0.705 &0.800 &0.754\\
				
				%\hline
				
				PrDiMP-50~\cite{PrDiMP} &CVPR2020&0.514 &0.461 &0.388 &0.608 &0.688 &0.598
				&0.704 &0.816 &0.758\\
				
				%\hline
				
				PrDiMP-18~\cite{PrDiMP} &CVPR2020&0.466 &0.421 &0.354 &- &0.645 &0.564
				&0.691 &0.803 &0.750\\
				
				%\hline
				
				DiMP-50~\cite{DiMPBhat2019LearningDM} &ICCV2019&0.521 &0.467 &0.395 &0.567 &0.650 &0.569
				&0.687 &0.801 &0.740\\
				
				%\hline
				
				DiMP-18~\cite{DiMPBhat2019LearningDM} &ICCV2019&0.487 &0.437 &0.369 &\multicolumn{1}{c}{-} &0.610 &0.532
				&0.666&0.785 &0.723\\
				
				%\hline
				
				ATOM~\cite{ATOM} &CVPR2019&0.480 &0.431 &0.360 &0.505 &0.576 &0.515
				&0.648 &0.771 &0.703\\
				
				\hline
		\end{tabular}}
	\end{center}
	\vspace{-0.3cm}
\end{table*}

As a typical MDNet-based RGBT tracking method, MANet shows a good tracking performance under the training of small-scale datasets. Therefore we also conduct the same retraining experiment on MANet. We can see that PR and SR scores on RGBT234 are also improved by retraining it using LasHeR \emph{training} set. The performance improvement of the two types of trackers based on MDNet and Siamese framework proves the importance of the large-scale high-diversity benchmark dataset for the training of deep trackers.

\subsection{Evaluation of RGB Trackers on LasHeR}

To further verify the diversity and challenge of our proposed LasHeR dataset, we present the results of some RGB trackers on LasHeR dataset (RGB videos), LaSOT~\cite{LaSOT} and TrackingNet~\cite{Mller2018TrackingNetAL} in Table~\ref{tab:RGB_results_lasher}, including TransT~\cite{TransT}, SiamFC++~\cite{SiamFC++},  PrDiMP-50~\cite{PrDiMP},PrDiMP-18~\cite{PrDiMP}, DiMP-50~\cite{DiMPBhat2019LearningDM},  DiMP-18~\cite{DiMPBhat2019LearningDM} and ATOM~\cite{ATOM}. From the results we can see that the performance on LasHeR is greatly lower than on LaSOT and TrackingNet, and it suggests that our LasHeR dataset is more challenging.

To show the performance of the state-of-the-art RGB tracking algorithms, we also add the above RGB trackers on the proposed LasHeR \emph{testing} set. The results are shown in Table~\ref{tab:RGB_results_lashertest}, and we can see that TransT performs best in RGB trackers. Note that the performance of RGB trackers performs well against most RGBT trackers. The reason is that these RGB trackers are trained on large-scale datasets while RGBT trackers are trained on small-scale ones. To show the complementary benefits of RGBT data, we compare the RGBT tracker mfDiMP~\cite{zhang2019multi} retrained by the large-scale RGBT tracking dataset LasHeR \emph{training} set with these RGB trackers. From the results in Table~\ref{tab:retraining} and Table~\ref{tab:RGB_results_lashertest}, we can observe that  mfDiMP greatly outperforms the second best tracker TransT. It fully demonstrates the effectiveness of RGBT data in visual tracking.

\begin{table}[htbp]\scriptsize
	\setlength{\abovecaptionskip}{-0.1cm}
	\caption{Evaluation results of  TransT~\cite{TransT}, SiamFC++~\cite{SiamFC++}, PrDiMP~\cite{PrDiMP}, DiMP~\cite{DiMPBhat2019LearningDM} and ATOM~\cite{ATOM} on LasHeR \emph{testing} set.}
	\label{tab:RGB_results_lashertest}
	\renewcommand\arraystretch{1.3}
	\begin{center}
		\setlength{\tabcolsep}{3mm}{
			\begin{tabular}{|c| c cc|}
				\hline
				
				Methods& PR & NPR & SR\\ 
				\hline
				
				%\textbf{mfDiMP~\cite{zhang2019multi}}&RGBT&0.599&0.467\\
				TransT~\cite{TransT}&0.529&0.487&0.395\\
				SiamFC++~\cite{SiamFC++}&0.348&0.308&0.274\\
				PrDiMP-50~\cite{PrDiMP}&0.444&0.392&0.337\\
				PrDiMP-18~\cite{PrDiMP}&0.397&0.357&0.303\\
				DiMP-50~\cite{DiMPBhat2019LearningDM}&0.442&0.385&0.336\\
				DiMP-18~\cite{DiMPBhat2019LearningDM}&0.415&0.363&0.318\\
				ATOM~\cite{ATOM}&0.406&0.355&0.307\\

				\hline
		\end{tabular}}
	\end{center}
	\vspace{-0.4cm}
\end{table}

\section{Conclusion}

In this paper, we propose LasHeR, which is, as far as we know, the largest dataset with high-quality dense bounding box annotations to date for RGBT tracking. By constructing such a large-scale dataset, we hope to alleviate the contradiction between the hunger for data of deep RGBT trackers and the inability of existing datasets to meet the training needs of deep trackers. At the same time, we also hope to propose LasHeR to make up for the limitations of the existing datasets that are far from the real-world tracking applications in terms of challenges and scenarios. Proposing such a large-scale multi-challenge dataset allows the tracker to get a comprehensive evaluation on the data closed to the real-world applications. In addition, we also conduct experiments to prove that our dataset is challenging and necessary, and there is still a lot of room for the research and development of RGBT tracking.
In the future, we plan to expand the data of categories to further enhance the diversity of the data set. In addition, we will complete the annotations with other information including target masks and semantic descriptions.

\bibliographystyle{IEEEtran}
\bibliography{egbib}

\begin{IEEEbiography}[{\includegraphics[width=1in,height=1.25in,clip,keepaspectratio]{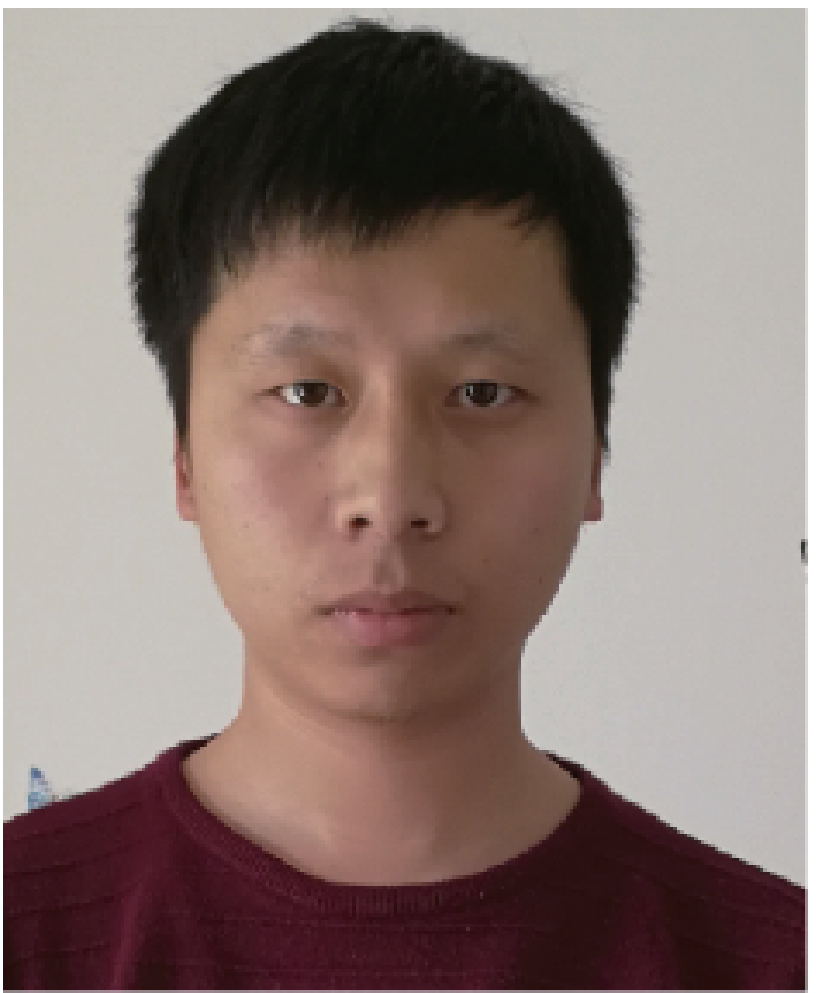}}]{Chenglong Li}
received the M.S. and Ph.D. degrees from the School of Computer Science and Technology, Anhui University, Hefei, China, in 2013 and 2016, respectively. He is currently an Associate Professor with the School of Artificial Intelligence, Anhui University. His research interests include computer vision and deep learning.
	\vspace{-10 mm}
\end{IEEEbiography}

\begin{IEEEbiography}[{\includegraphics[width=1in,height=1.25in,clip,keepaspectratio]{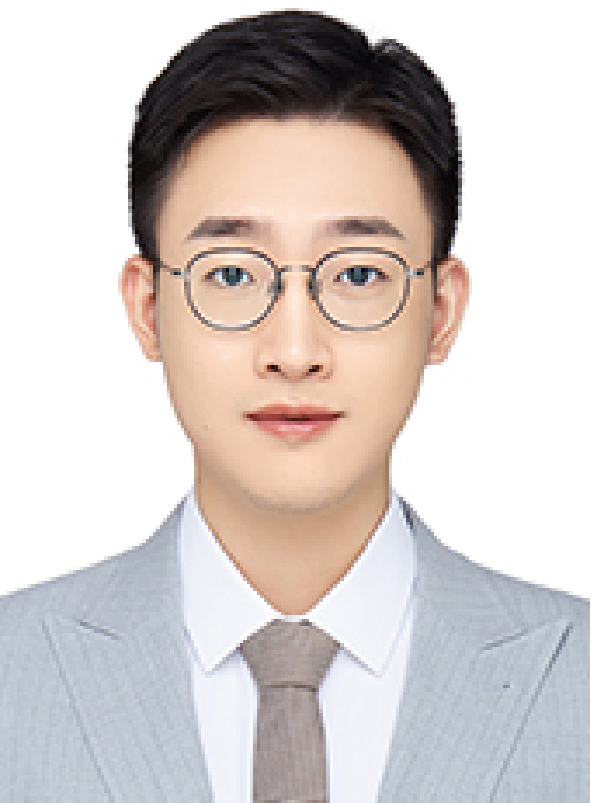}}]{Wanlin Xue}
received the B.Eng. degree from  Anhui University,Hefei,China,in 2019. He is currently pursuing the M.Eng degree in Computer Science and Technology from Anhui University, Hefei, China.  His current research interests include computer vision and deep learning and now is focusing on the research of RGBT object tracking.
\end{IEEEbiography}

\begin{IEEEbiography}[{\includegraphics[width=1in,height=1.25in,clip,keepaspectratio]{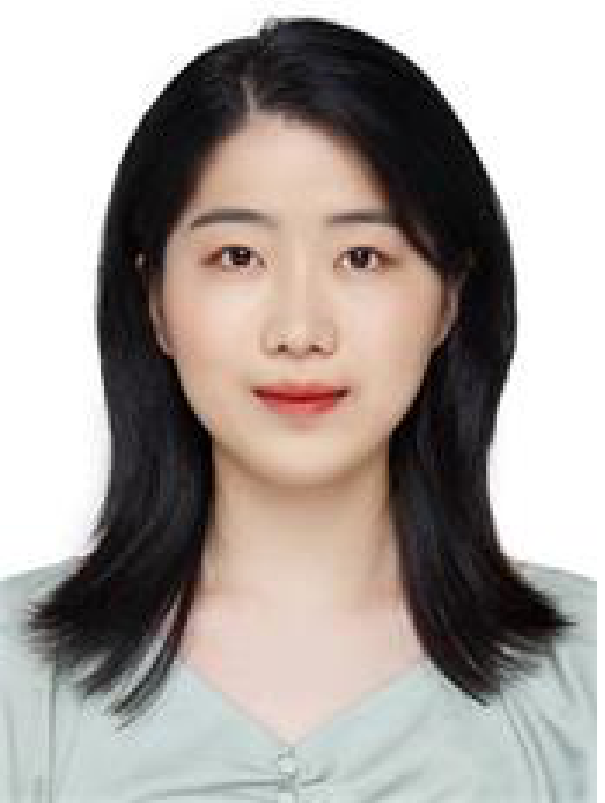}}]{Yaqing Jia}
received the B.Eng. degree from College Of Economy and Technology Of Anhui Agricultural University , Hefei, China, in 2019. She is currently pursuing the M.Eng degree with  Anhui University ,Hefei,China. Her current research interests include computer vision and deep learning and now is focusing on the research of 3D thermal infrared reconstruction.
\end{IEEEbiography}

\begin{IEEEbiography}[{\includegraphics[width=1in,height=1.25in,clip,keepaspectratio]{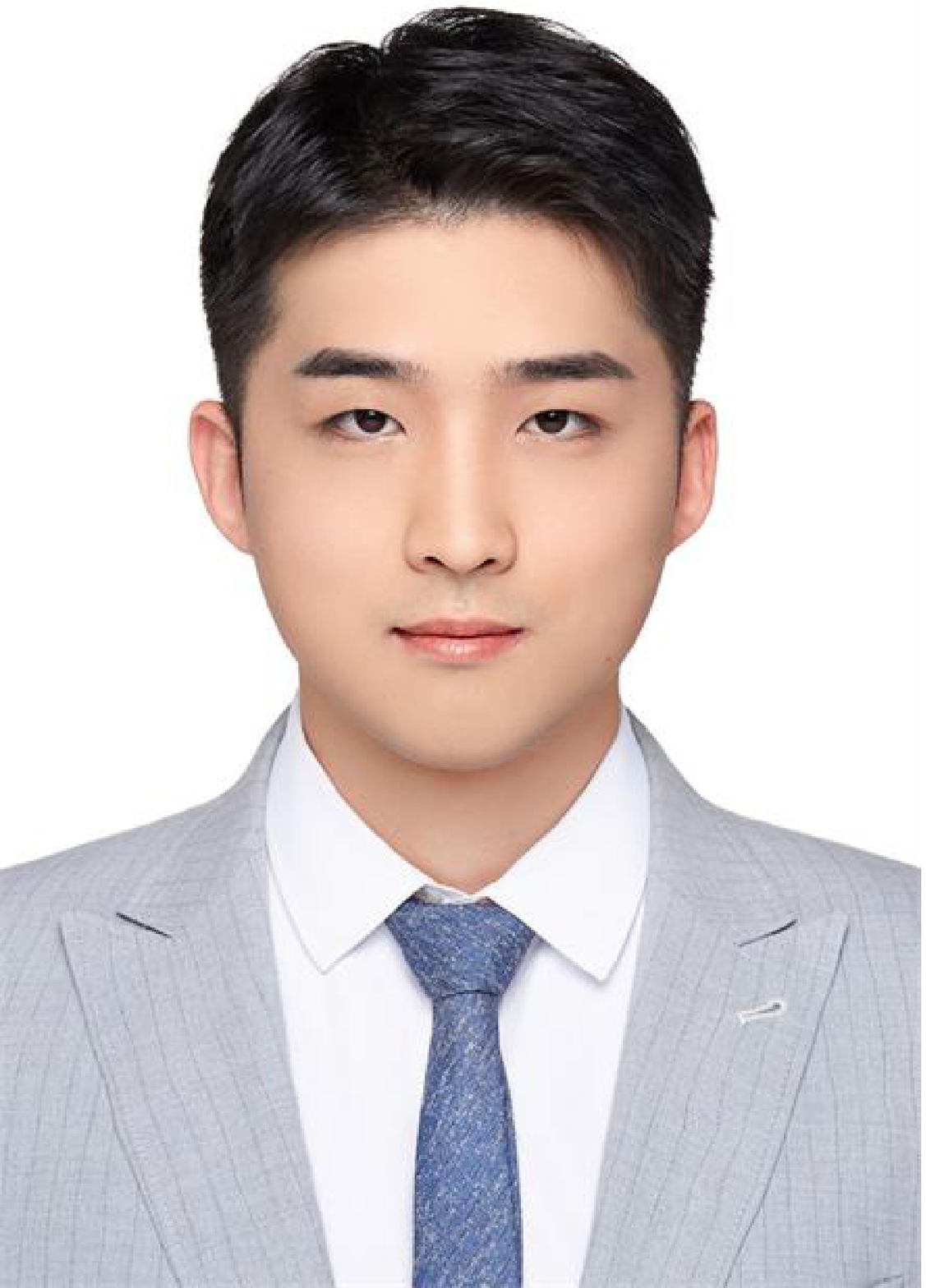}}]{Zhichen Qu}
is pursuing B.Eng. degree in Anhui University, Hefei, China. His current research interests include computer vision and deep learning and now is focusing on the research of vehicle detection based on compact bounding box.
\end{IEEEbiography}

\begin{IEEEbiography}[{\includegraphics[width=1in,height=1.25in,clip,keepaspectratio]{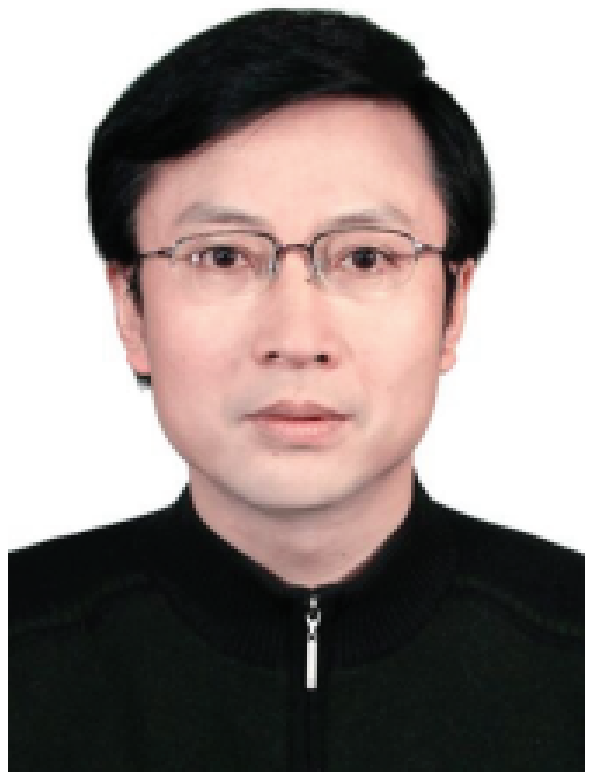}}]{Bin Luo}	
received the B.Eng. degree in electronics and the M.Eng. degree in computer science from Anhui University, Hefei, China, in 1984 and 1991, respectively, and the Ph.D. degree in computer science from the University of York, York, U.K., in 2002. He is currently a Professor with Anhui University. His current research interests include graph spectral analysis, large image database retrieval, image and graph matching and statistical pattern recognition.
\end{IEEEbiography}

\begin{IEEEbiography}[{\includegraphics[width=1in,height=1.25in,clip,keepaspectratio]{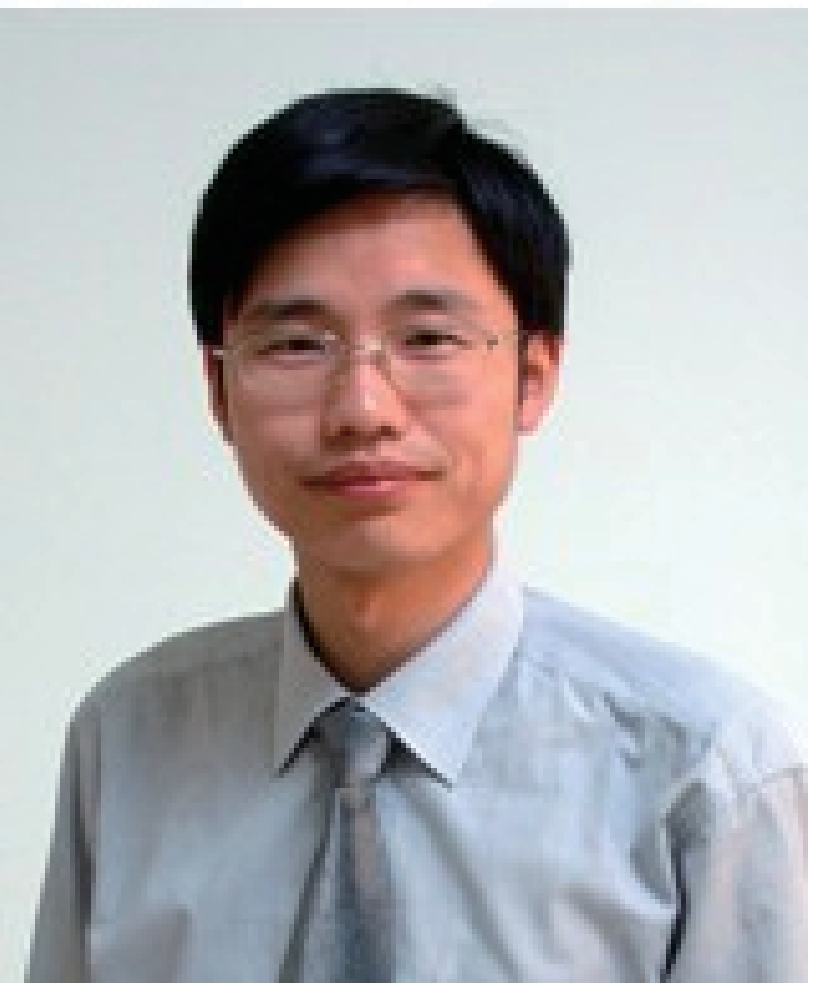}}]{Jin Tang}	
received the B.Eng. degree in automation and the Ph.D. degree in computer science from Anhui University, Hefei, China, in 1999 and 2007,
	respectively. He is currently a Professor with the School of Computer Science and Technology, Anhui University. His current research interests include computer vision, pattern recognition, machine learning, and deep learning.
\end{IEEEbiography}

\begin{IEEEbiography}[{\includegraphics[width=1in,height=1.25in,clip,keepaspectratio]{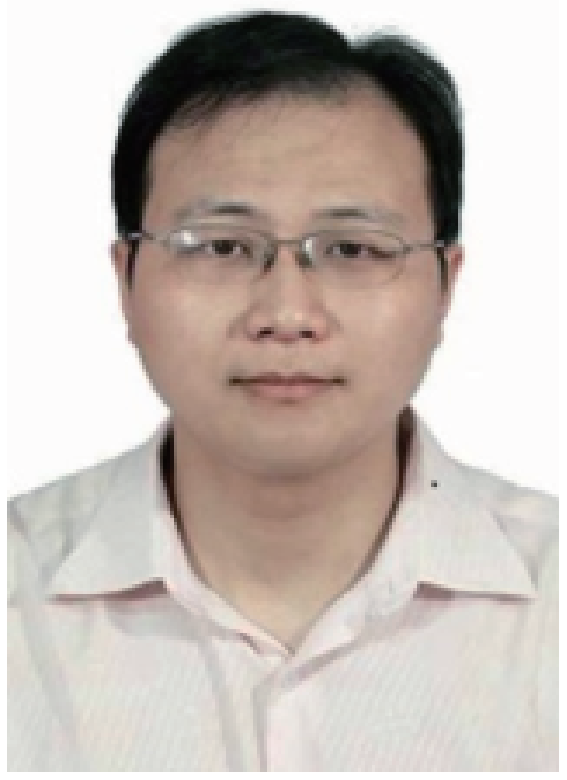}}]{Dengdi Sun}	
received the B.Eng. degree in statistics, the M.Eng. degree in mathematics and the Ph.D. degree in computer science from Anhui University, Hefei, China, in 2005, 2008 and 2012, respectively. Currently, he is an associate professor in the School of Computer Science and Technology, Anhui University. His main research interests include computer vision, machine learning and deep learning.
\end{IEEEbiography}

\end{document}